%% file: acl_latex.tex
\pdfoutput=1

\documentclass[11pt]{article}

\usepackage[preprint]{acl}

\usepackage{times}
\usepackage{latexsym}

\usepackage[T1]{fontenc}

\usepackage[utf8]{inputenc}

\usepackage{microtype}

\usepackage{inconsolata}

\usepackage{graphicx}

\usepackage{algpseudocode,algorithm}
\usepackage{hyperref}
\usepackage{multirow}
\usepackage{graphicx}
\usepackage{pdfpages}
\usepackage{url}
\usepackage{xcolor}
\usepackage{epsfig}
\usepackage{adjustbox}
\usepackage{amsfonts}
\usepackage{amsmath}
\usepackage{amssymb}
\usepackage{booktabs} 
\usepackage{comment}
\usepackage{caption}
\usepackage{subcaption}
\usepackage{textcomp}
\usepackage{relsize}
\usepackage{stmaryrd}
\usepackage{bbm}
\usepackage{rotating}
\usepackage{helvet}
\usepackage{courier}
\usepackage{natbib}
\usepackage{cleveref}
\usepackage{amsmath}
\usepackage{tabularx}
\usepackage{multirow}
\usepackage{multicol}
\usepackage[normalem]{ulem}
\usepackage{cleveref}
\usepackage{arydshln}

\usepackage{xspace}
\usepackage{graphicx}
\usepackage{caption}
\usepackage{multirow}
\usepackage{multicol}
\usepackage{makecell}
\usepackage{booktabs}
\usepackage{tabularx}
\usepackage{adjustbox}
\usepackage{pifont}
\usepackage{soul}
\usepackage{enumitem}
\usepackage{amsmath}
\usepackage{amssymb}
\usepackage{comment}
\usepackage{enumitem}
\usepackage{cleveref}
\usepackage{arydshln}
\usepackage{graphicx,dblfloatfix}
\usepackage{enumitem}
\usepackage{verbatim}
\usepackage{breakurl}

\input{macro}

\title{\cactus{}~\textsc{Cactus}: Towards Psychological Counseling Conversations using Cognitive Behavioral Theory}

\author{
  Suyeon Lee$^{1}$\thanks{~~Equal contribution} \qquad \textbf{Sunghwan Kim}$^{1}$\footnotemark[1] \qquad  
  \textbf{Minju Kim}$^{1}$\footnotemark[1] \qquad \\
  \textbf{Dongjin Kang}$^{2}$ \quad
  \textbf{Dongil Yang}$^{1}$ \quad
  \textbf{Harim Kim}$^{3}$ \quad \\
  \textbf{Minseok Kang}$^{2}$ \quad 
  \textbf{Dayi Jung}$^{3}$ \quad 
  \textbf{Min Hee Kim}$^{3}$ \quad 
  \textbf{Seungbeen Lee}$^{1}$ \quad \\ 
   \textbf{Kyoung-Mee Chung}$^{3}$\quad 
   \textbf{Youngjae Yu}$^{1}$\quad
   \textbf{Dongha Lee}$^{1}$ \quad \textbf{Jinyoung Yeo}$^{1}$ \\ 
Department of Artificial Intelligence$^{1}$, Computer Science$^{2}$, Psychology$^{3}$\\ Yonsei University \\
  \texttt{\{isuy.groot,kimsh8564,minnju,jinyeo\}@yonsei.ac.kr}\\
}

\begin{document}
\maketitle
\input{latex/0-abstract}
\input{latex/1-introduction}
\input{latex/2-design-consideration}

\input{latex/3-method}

\input{latex/4-experiments}

\input{latex/5-related-work}
\input{latex/6-conclusion}

\bibliography{custom}

\appendix
\input{latex/9-appendix}

\end{document}

%% file: macro.tex
\crefformat{section}{\S#2#1#3}
\crefformat{subsection}{\S#2#1#3}
\crefformat{subsubsection}{\S#2#1#3}
\crefrangeformat{section}{\S\S#3#1#4 to~#5#2#6}
\crefmultiformat{section}{\S\S#2#1#3}{ and~#2#1#3}{, #2#1#3}{ and~#2#1#3}

\newcommand{\eg}{{\it e.g.}}%
\newcommand{\ie}{{\it i.e.}}%

\newcommand{\ourdatas}{\textsc{Cactus}\xspace}

\definecolor{yellow-green}{rgb}{0.3, 0.5, 0.0}

\definecolor{lightergray}{RGB}{230,230,230}
\definecolor{DarkGreen}{RGB}{30,130,30}
\newcommand{\cmark}{\textcolor{DarkGreen}{\ding{51}}}
\newcommand{\xmark}{\textcolor{red}{\ding{55}}}
\newcommand{\cxmark}{\raisebox{-2pt}{\includegraphics[width=1em]{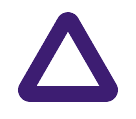}}}
\newcommand\cactus{\raisebox{-2pt}{\includegraphics[width=1em]{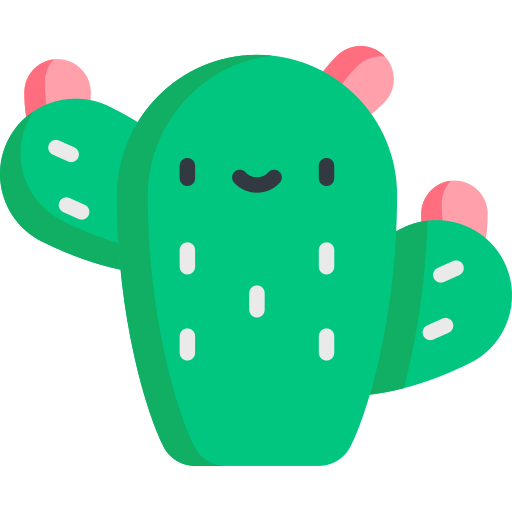}}}

\definecolor{lightblue}{RGB}{224,236,247}
\definecolor{deepblue}{RGB}{9,46,107}

%% file: latex/0-abstract.tex
\begin{abstract}
Recently, the demand for psychological counseling has significantly increased as more individuals express concerns about their mental health. This surge has accelerated efforts to improve the accessibility of counseling by using large language models (LLMs) as counselors. To ensure client privacy, training open-source LLMs is necessary, but this approach faces a key challenge: the absence of realistic counseling datasets. To address the issue of dataset scarcity, we introduce \textsc{Cactus}, a multi-turn dialogue dataset that emulates real-life interactions using the goal-oriented and structured approach of Cognitive Behavioral Therapy (CBT).
We create a diverse and realistic dataset by (1) designing clients with varied, specific personas, and (2) having counselors systematically apply CBT techniques in their interactions. To assess the quality of our data, we benchmark against established psychological criteria used to evaluate real counseling sessions, ensuring alignment with human expert evaluations.
Experimental results demonstrate that \textsc{Camel}, a model trained with \textsc{Cactus}, outperforms other models in counseling skills, highlighting its effectiveness and potential as a counseling agent.
We make our dataset, model, and code publicly available.\footnote{{\href{https://github.com/coding-groot/cactus}{https://github.com/coding-groot/cactus}}}
\end{abstract}

%% file: latex/1-introduction.tex
\section{Introduction}
According to cognitive therapy, psychological problems arise when individuals irrationally interpret external events~\citep{Powles1974BeckAT}. Therefore, the goal of a counselor is to identify and correct irrational and negative thinking patterns of the client through conversation, guiding clients to change their negative thoughts to positive thoughts (\ie, reframing thoughts)~\citep{beck2020cognitive}. Cognitive Behavioral Therapy (CBT) uses reframing techniques tailored to the client’s characteristics, making it one of the most widely used counseling strategies~\citep{greimel2011cognitive}.

\input{figures/intro_figure}
For example, consider a client who thinks, ``\textit{Many consider me a nerd. I'm such a useless social person}'' as depicted in Figure~\ref{fig:motivation}. The client describes themselves negatively as a nerd and a useless social person, leading to an overly negative self-assessment. These unhelpful thought patterns are classified as \textit{labeling}, contributing to their distorted self-perception. To reframe the negative thoughts of the client, the counselor can apply the CBT technique of \textit{Alternative Perspective.} Using \textit{Alternative Perspective} helps the client move beyond these negative labels to develop a more accurate and positive self-view.

Recently, there has been growing interest in using closed-source models (\eg, ChatGPT) as counselors~\citep{Raile2024chatgpt_psychotherapists, berrezueta2024future_adhd} due to their remarkable conversational ability~\citep{pirnay2023chatgpt}. However, significant data privacy concerns emerge as these approaches send conversation data, which often contains highly personal information, to third-party providers. To address data privacy concerns, it is crucial to utilize an open-source model~\citep{hicke2023aita}, where performance can be ensured by training on datasets that closely resemble real-world data. While using actual counseling data would be ideal, such dataset is not publicly available due to ethical concerns, especially the risk of individual identification~\citep{qiu2023smile}. As a result, synthetic datasets that closely emulate real-world counseling scenarios present an alternative to address this issue.

Prior attempts to create synthetic counseling datasets have mainly focused on single-turn counseling strategies, significantly diverging from real-world counseling practices~\citep{sharma2023cognitive, sun-etal-2021-psyqa, liu2023chatcounselor}. In response, recent efforts have aimed to implement multi-turn strategies~\citep{xiao2024healme, qiu2023smile}. Nevertheless, these approaches often involve only 2-3 turns of dialogue or convert single-turn interactions into multi-turn conversations. Therefore, there is a need for a more realistic approach to psychological counseling that ensures therapeutic consistency and progression across multiple turns.

To tackle these challenges, we introduce \textsc{Cactus} (\textbf{\underline{C}}BT-\textbf{\underline{a}}ugmented \textbf{\underline{C}}ounseling Cha\textbf{\underline{t}} Corp\textbf{\underline{us}}), a publicly available multi-turn realistic counseling dataset. By carefully guiding LLMs to simulate interactions between counselors and clients, \textsc{Cactus} captures the depth and flow of psychological counseling. Human evaluation results demonstrate that \textsc{Cactus} surpasses the existing counseling datasets in terms of helpfulness and empathy.

An assessment of a counselor's abilities needs to consider both their communication skills and the client's psychological changes before and after the counseling session. However, existing methods, such as automatic or single-turn evaluations, are not suitable for assessing a counselor's abilities~\citep{smith-etal-2022-human, liu2023chatcounselor}.
To this end, we introduce \textsc{CounselingEval}, an evaluation framework that assesses the ability of counselor in the perspective of both counselor and client within multi-turn counseling dialogues.
In \textsc{CounselingEval}, we assume AI clients, which interact with the counselor (\ie, target for evaluation), using a set of client information.
Based on the multi-turn conversations between these virtual clients and the counselor, we evaluate the counseling conversations through a modified version of real-world evaluation methods including CTRS and PANAS~\citep{goldberg2020structure, saxon2017affective}.
The results of \textsc{CounselingEval} present the effectiveness of \textsc{Cactus} through both LLM-based evaluation and psychological expert evaluation.
Furthermore, our model \textsc{Camel}, trained on \textsc{Cactus}, also outperforms other baseline models in counseling abilities, suggesting that \textsc{Cactus} serves as a valuable resource for enhancing the psychological abilities of open-source LLMs.

%% file: figures/intro_figure.tex
\begin{figure}[t!]
    \centering
    \includegraphics[width=0.98\linewidth]{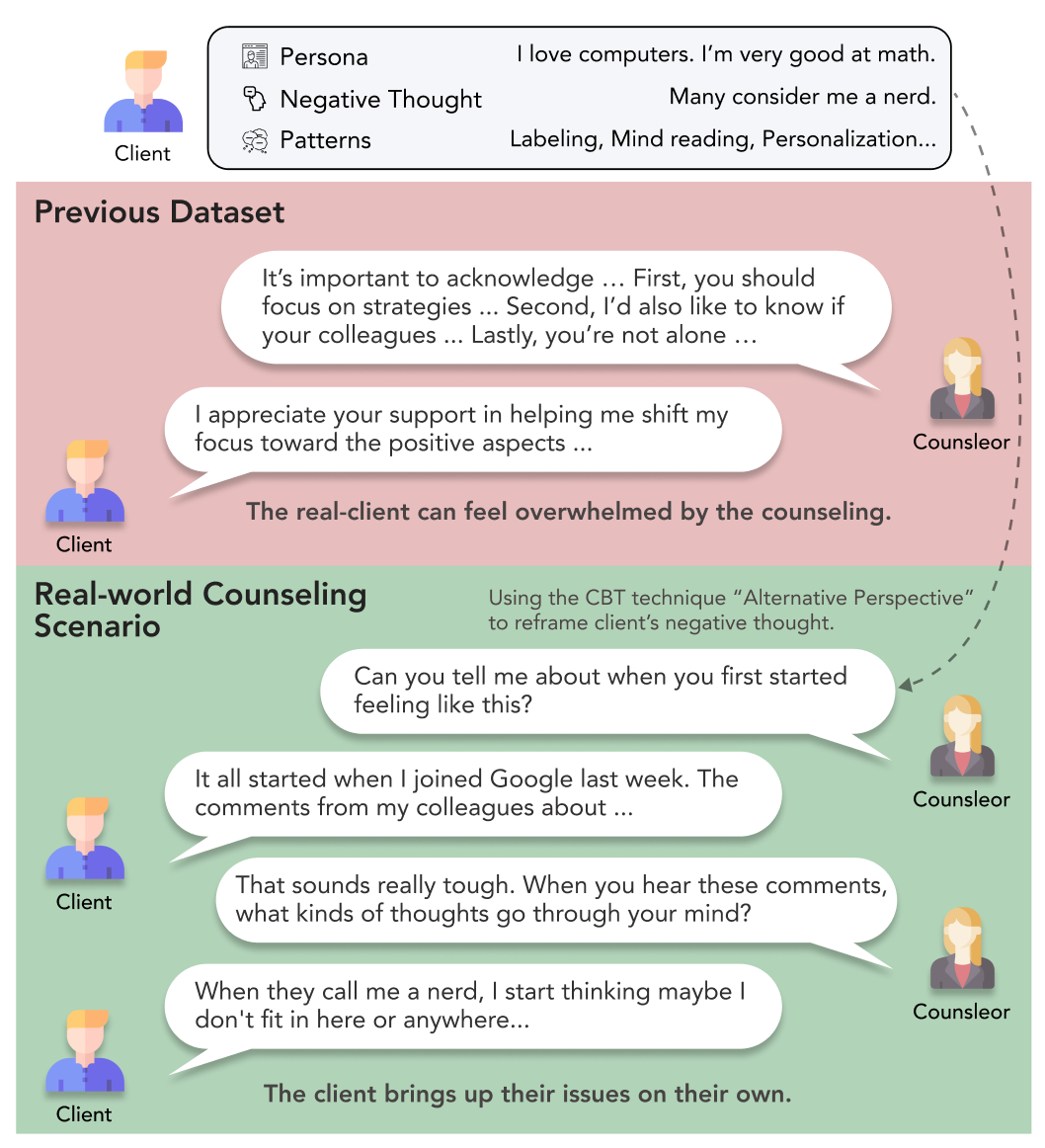}
    \caption{Comparing a previous counseling dataset with real-world scenarios: The dataset shows counselors conveying large amounts of information in a single turn, whereas real-world counseling involves active, collaborative communication between a counselor and client.}
    \label{fig:motivation}
\end{figure}

%% file: latex/2-design-consideration.tex
\section{Design Considerations for \textsc{Cactus}}
\label{sec:design}
We aim to develop a counseling dataset simulating real-world scenarios by simulating conversations between an AI counselor and client~\citep{xiao2024healme, zhou2023sotopia}. Before the conversation begins, we assume an initial intake session, where basic client information (\eg, name, age, reasons for seeking therapy) is shared, has already been conducted. During the conversation, the AI counselor guides clients through reframing their thoughts using CBT techniques, while personas from PatternReframe~\citep{maddela2023training} help simulate AI clients. In the following sections, we discuss the challenges when utilizing LLMs in counseling dataset generation.\footnote{In this section, we utilize \texttt{gpt-3.5-turbo-0125} and \texttt{gpt-4o-2024-05-13}.}

\input{figures/cbt_tech_figure}

\subsection{Can LLMs be Competent Counselors?}
\label{ssec:test}
While LLMs are known to possess knowledge of psychological therapy concepts such as CBT techniques~\citep{OpenAICommunity}, there are still shortcomings in using LLMs for counseling. 
 
\paragraph{AI counselors have limitations in selecting CBT techniques.}
We conduct an experiment to compare the selection of CBT techniques between LLMs and human experts. We provide psychological experts with 150 randomly selected client thoughts and patterns from PatternReframe to annotate appropriate CBT techniques. 
In Figure~\ref{fig:cbt_tech_distribution}, GPT-4o exhibits a biased selection of CBT techniques, with \textit{Evidence-Based Questioning} being utilized in almost half of all cases. However, the selection by experts is found to be more evenly distributed. This finding reveals that each LLM has its own bias in selecting techniques, as also pointed out in \citet{kang2024can}. Additionally, it demonstrates that a discrepancy between LLMs and human experts exists. Further details are in Appendix~\ref{app:desing_consider1}.

\paragraph{AI counselors tend to suggest direct reframing of the clients' thoughts.}
The role of a counselor is not to provide instant solutions but to help the client discover their own solutions through fostering a collaborative relationship. However, as shown in Figure~\ref{fig:chatgpt_shortage}, AI counselors often present reframed thoughts directly, such as \textit{``Remind yourself of positive moments.''} This approach may encounter client resistance and hinder clients from independently reframing negative thoughts.

\input{figures/chatgpt_shortage_figure}
\subsection{Can LLMs act like Real Clients?}
\label{sec:design_client}
We simulate an AI client with diverse behaviors by providing it with information such as persona, negative thoughts, and patterns before initiating the conversation, as shown in Figure~\ref{fig:chatgpt_shortage}. AI clients encounter challenges in emulating human-like interaction, which hinders the generation of authentic and high-quality conversational data.

\paragraph{AI clients tend to express the provided information explicitly.}
In Figure~\ref{fig:chatgpt_shortage}, the client states, \textit{``It might be labeling,''} using psychological terms to explain their situation. It differs from real-life clients, who describe their reasons for seeking help with detailed and contextual stories without using psychological terminology.

\paragraph{AI clients tend to be overly positive.}
AI clients consistently exhibit a positive attitude during counseling sessions~\citep{serapiogarcía2023personality}, which often results in compliant responses such as \textit{``I am willing to try it.''} AI clients lack the diverse range of attitudes (\eg, strongly negative attitudes) that real-life clients often express.

%% file: figures/cbt_tech_figure.tex
\begin{figure}[h!]
    \centering
    \includegraphics[width=0.8\linewidth]{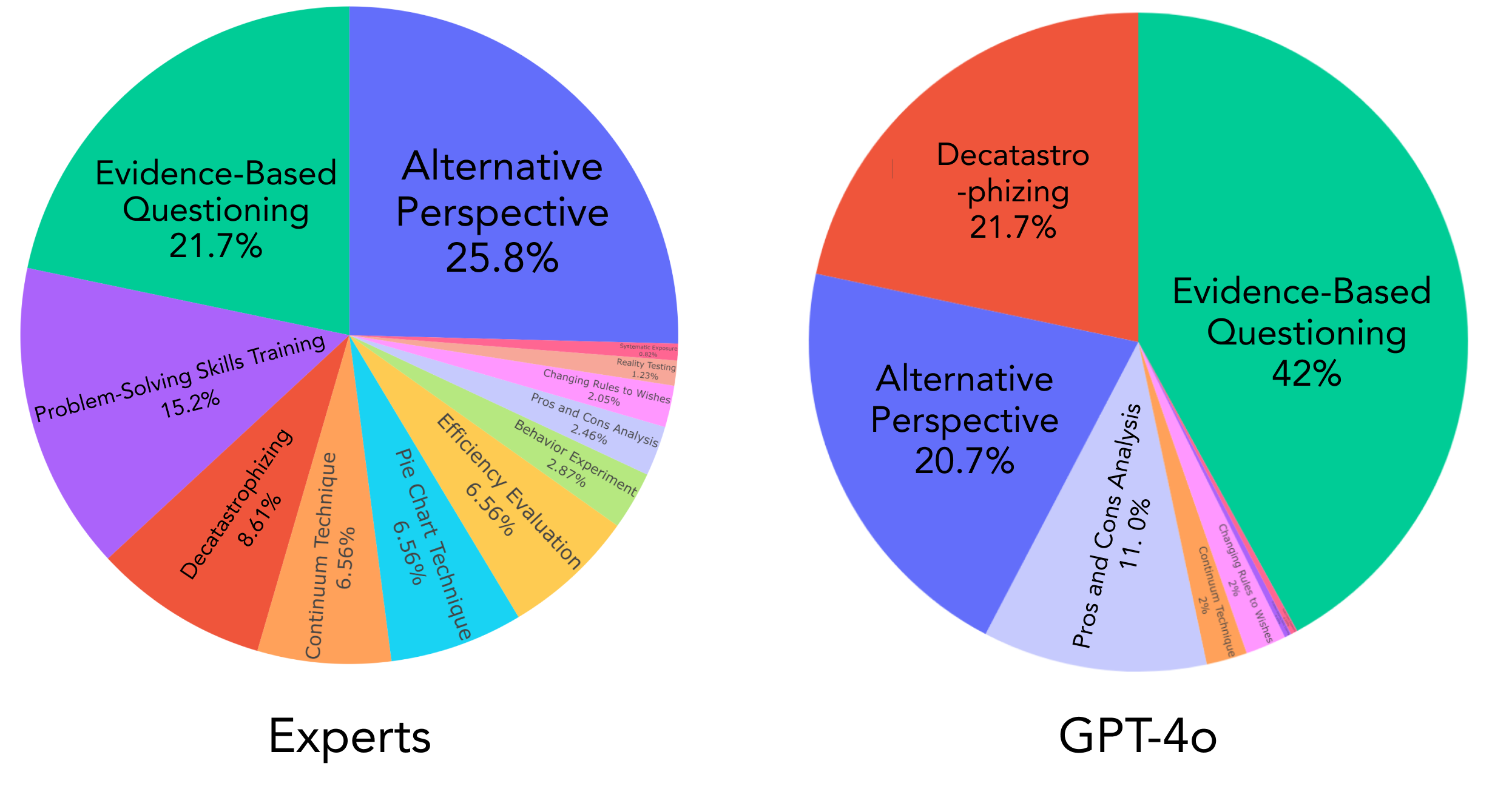}
    \caption{Comparison of the distribution of CBT techniques selected by GPT-4o and psychological experts. The results of GPT-3.5-Turbo are shown in Figure~\ref{fig:cbt_tech_distribution_chatgpt}.
    }
    \label{fig:cbt_tech_distribution}
    \vspace{-10pt}
\end{figure}

%% file: figures/chatgpt_shortage_figure.tex
\begin{figure}[t!]
    \centering
    \includegraphics[width=0.94\linewidth]{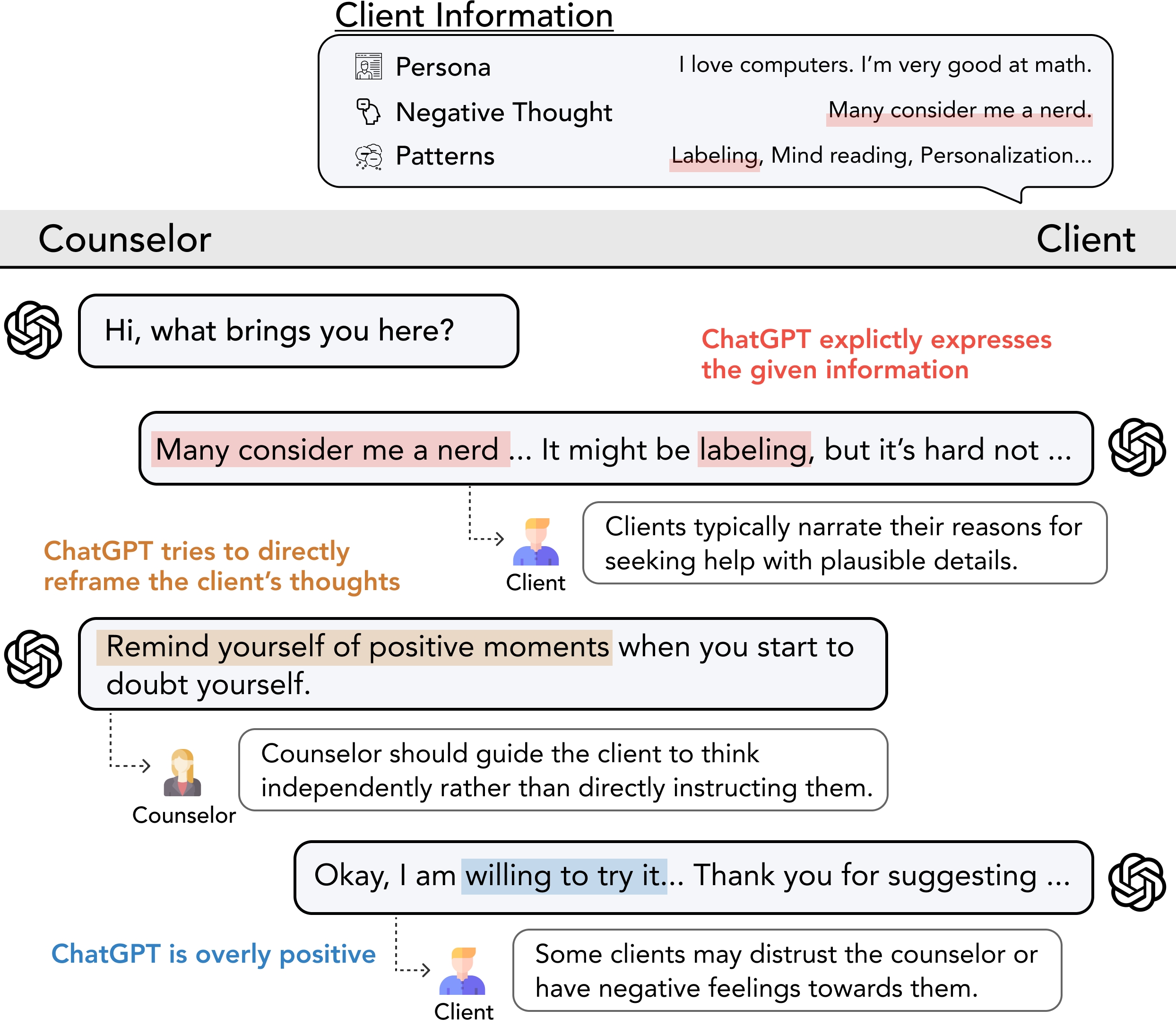}
    \caption{Empirical investigations into the problems of using ChatGPT as an AI counselor and AI client. Details of experiments are in Appendix~\ref{app:desing_consider2}.}
    \label{fig:chatgpt_shortage}
\end{figure}

%% file: latex/3-method.tex
\input{tables/data_comparison}
\section{\cactus{}~\textsc{Cactus}: A Psychological Counseling Dataset using CBT}
We describe the dataset construction process of \textsc{Cactus}, a psychological counseling dataset that can also be used for training a counselor agent. We carefully design the dataset construction process to overcome the LLMs' challenges presented in Section~\ref{sec:design}. In this section, we outline the generation of dialogues with LLMs and the filtering process for realistic and specialized data. The overview of our approach is in Figure~\ref{fig:challenge}  and an example of \ourdatas{} is in Appendix~\ref{app:details_of_cactus}.
\input{figures/challenge_figure}
\subsection{Dataset Construction}
\paragraph{Source dataset.}
To generate counseling dialogues, we use personas, thoughts, and patterns from the PatternReframe~\citep{maddela2023training} as contexts for simulating clients. We choose PatternReframe for two reasons: (1) the thoughts closely mirror those of actual clients, and (2) individuals with psychology backgrounds contributed to its creation through crowd-sourcing. Further details are provided in Appendix~\ref{app:pattern_filtering}.

\paragraph{Counselor simulation.}
The counselor simulation aims to enhance the effectiveness of counseling in multi-turn interactions. Initially, counselors receive the client's negative thought, patterns, and reframed thoughts from PatternReframe. Negative thoughts and patterns refer to the client's psychological issues, while reframed thoughts are the result of transforming those negative thought patterns. The simulated counselors' task is to select the top three CBT techniques that best support this reframing process, a fundamental goal of counseling. To ensure systematic counseling, counselors undergo a planning process that incorporates CBT techniques before sessions. Through the planning process, counselors effectively guide the clients toward the independent discovery of solutions rather than directly suggesting reframed thoughts.

\paragraph{Client simulation.}
To address the limitations of AI clients (\cref{sec:design_client}), we convert negative thoughts into detailed client narratives~\citep{radfordlanguage}. Inspired by \citet{schmidgall2024agentclinic}, which simulates patient agents with templates for demographics, clinical history, and symptoms, we prompt LLMs to fill out an intake form for client modeling. An intake form includes information on the reason for seeking help, as well as details such as name, gender, and age. This process allows us to simulate specific clients and facilitate effective counseling sessions. 
Furthermore, to simulate a realistic and diverse counseling environment, we establish three distinct client attitudes (\ie, positive, neutral, and negative) reflecting the variability in client behaviors. Following \citet{li2023understanding}, which categorized clients' attitudes into positive and negative, each with five and six fine-grained behaviors respectively, we assign specific behaviors such as providing information or self-criticism to each attitude. More details are in Appendix~\ref{app/client_simulation}.

\paragraph{Dialogue generation process.}
There are two methods for generating counseling dialogues: (1) assigning the roles of client and counselor to different models (\ie, \textit{two-agent mode}); (2) providing the information of client and counselor to generate in a script (\ie, \textit{script mode}).
First, we conduct experiments comparing the two-agent mode~\citep{zhou2023sotopia} and the script mode to determine which method generates more natural dialogue. As shown in Figure~\ref{fig:two_agent_script}, our findings indicate that the dialogue generated using the script mode is more natural and well-constructed compared to Two-Agent mode. Details of the experiment are provided in Appendix~\ref{app/dialogue_generation}. 
These findings align with those of \citet{zhou2024real}, further supporting our decision to utilize script mode for generating dialogues in counselor and client simulations using GPT-4o.
Recognizing that typical counseling sessions last 30-60 minutes, we generate longer dialogues to better reflect real counseling and provide more comprehensive interactions, as shown in Table~\ref{tab:dataset_comparison}.
\input{figures/two_script}
\input{figures/data_quality_human_eval}
\input{figures/data_quality_ablation}

\paragraph{Filtering.}
To ensure the quality of \textsc{Cactus}, we initially filter out dialogues exhibiting abnormal formats or an insufficient number of turns. Subsequently, we utilize the Cognitive Therapy Rating Scale (CTRS), a real-world metric used to assess the quality of CBT-based counseling, for dataset filtering~\citep{beck2020cognitive}. We select three items each from CTRS to evaluate general counseling skills and CBT-specific skills, with each criterion scoring between 0 and 6 points. Dialogues with an average score below 5 points for the 6 CTRS criteria are filtered out. Following this filtering process, 86.3\% of the initial conversations remain, constituting the 31,577 conversations in \textsc{Cactus}. More details about filtering are in Appendix~\ref{app:filtering}.

\subsection{Analysis of \textsc{Cactus}}

\paragraph{Large-scale and diverse content.}
The \textsc{Cactus} stands out for its large scale and diverse content, providing a comprehensive resource for training counselor agents. It consists of 31,564 dialogues with approximately 1 million utterances (Table~\ref{tab:dataset_comparison}). To ensure a broad spectrum of client personas, we incorporate diverse backgrounds such as the reason for therapy, age, and occupation (Appendix~\ref{app:analysis_cactus}). Additionally, we construct the dataset by considering three different attitudes identified in previous research, as well as the corresponding behaviors clients might exhibit during counseling~\citep{li2023understanding}.
This diversity enhances the realism and applicability of the dataset, making it a resource for developing robust counselor agents.

\paragraph{High quality.}
To assess the relative quality of the dataset, we conduct human evaluations on Amazon Mechanical Turk (AMT), comparing \textsc{Cactus} with Psych8k and SmileChat. We randomly sample 100 dialogues from each dataset and evaluate them according to four criteria: (1) Helpfulness, (2) Coherence, (3) Empathy, and (4) Guidance. 
Further details are in Appendix~\ref{app:analysis_cactus}.

Despite possessing a fewer instance compared to SmileChat and being synthetically generated unlike Psych8k, which utilizes real counseling conversations, Figure~\ref{fig:data_quality_human_eval} demonstrates that \textsc{Cactus} consistently outperforms both datasets across all evaluated metrics.
As shown in Figure~\ref{fig:data_quality_ablation}, \textsc{Cactus} shows the large performance gap in both helpfulness and empathy, highlighting the effectiveness of planning with CBT techniques for counseling.

%% file: tables/data_comparison.tex
\begin{table*}[!t]
    \centering
    \resizebox{0.92\linewidth}{!}{
    \begin{tabular}{lcccccc}
        \toprule
        & Planning & Publicly Available & Language  & $\#$ of dialogues & $\#$ of utterances &  $\#$ Avg. turns\\
        \midrule
        Psych8k~\citep{liu2023chatcounselor}  & \xmark & \cxmark & English & 8,187 & 16,374 & 1.0\\
        SmileChat~\citep{qiu2023smile}  & \xmark & \cmark & Chinese & 55,165 & 1,833,856 & 10.4 \\
        HealMe~\citep{xiao2024healme} & \cmark & \xmark & English & 1,300 & 7,800 & 3.0 \\
        CBT-LLM~\citep{na2024cbt} & \cmark & \xmark & Chinese &  22,327 & 44,654 & 1.0 \\
        \textsc{Cactus} & \cmark & \cmark & English & 31,577 & 995,512 & 16.6 \\
         \bottomrule
    \end{tabular}
    }
    \caption{
    {A comparison of \textsc{Cactus} with other psychological counseling datasets. The symbol \cxmark indicates conditional access, meaning the dataset can only be used with permission.}
    }
    \label{tab:dataset_comparison}
\end{table*}

%% file: figures/challenge_figure.tex
\begin{figure}[t!]
    \centering
    \includegraphics[width=\linewidth]{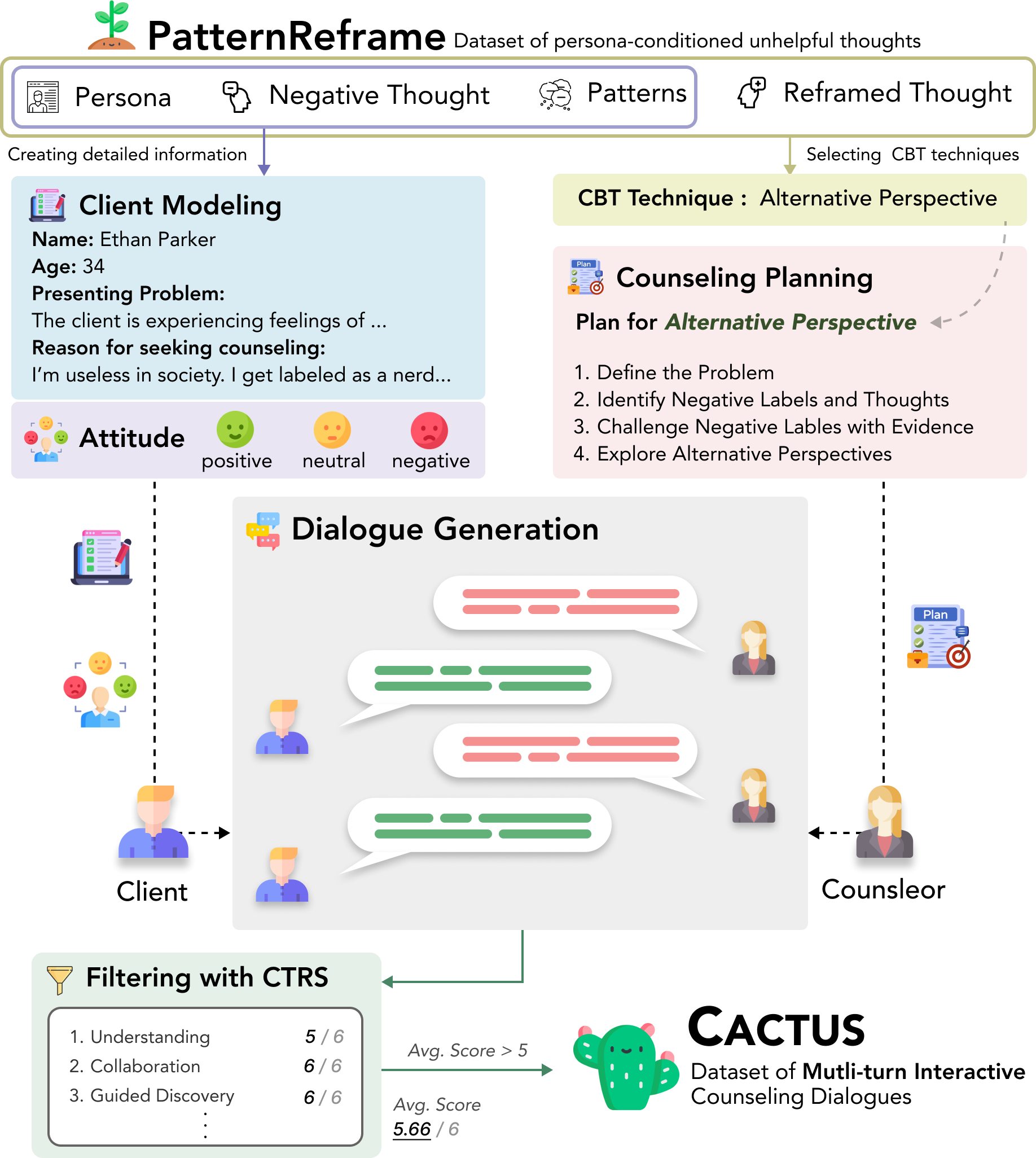}
    \caption{The overview of the data collection process of \cactus{} \textsc{Cactus}.}
    \label{fig:challenge}
    \vspace{-10pt}
\end{figure}

%% file: figures/two_script.tex
\begin{figure}[t!]
    \centering
    \includegraphics[width=\linewidth]{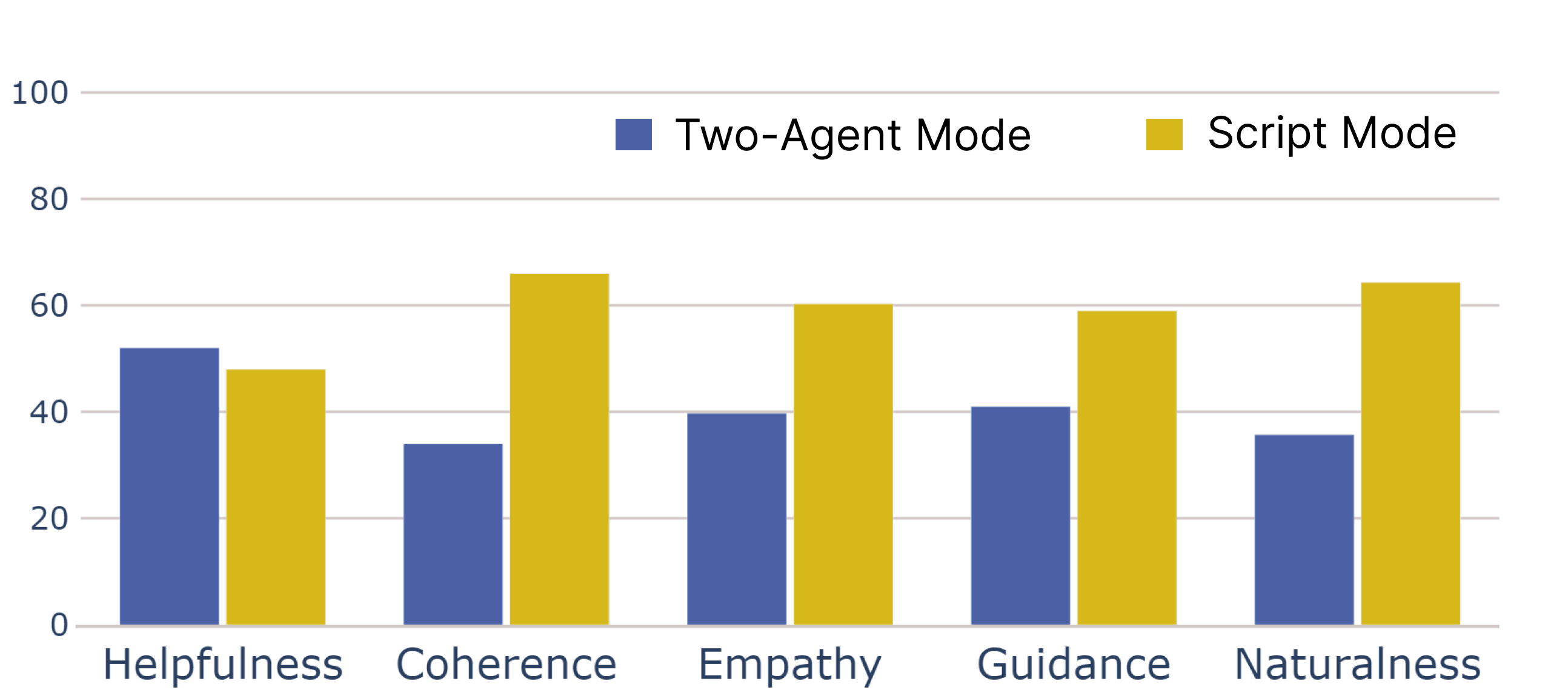}
    \caption{Comparision of Two-Agent mode and Script mode.}
    \label{fig:two_agent_script}
\end{figure}

%% file: figures/data_quality_human_eval.tex
\begin{figure*}[t!]
    \centering
    \includegraphics[width=0.92\linewidth]{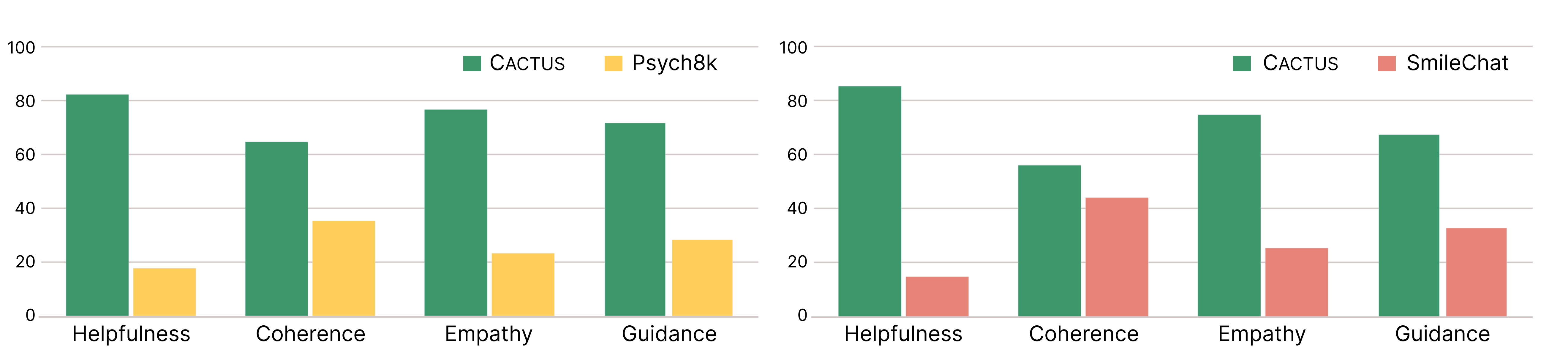}
    \caption{Results of head-to-head comparison between dialogues from \textsc{Cactus}, Psych8k~\citep{liu2023chatcounselor}, and SmileChat~\citep{qiu2023smile} based on human judgments. All results demonstrate statistically significant differences with $p<0.05$, except for the \textit{Coherence} between \textsc{Cactus} and SmileChat.}
    \label{fig:data_quality_human_eval}
\end{figure*}

%% file: figures/data_quality_ablation.tex
\begin{figure}[t!]
    \centering
    \includegraphics[width=\linewidth]{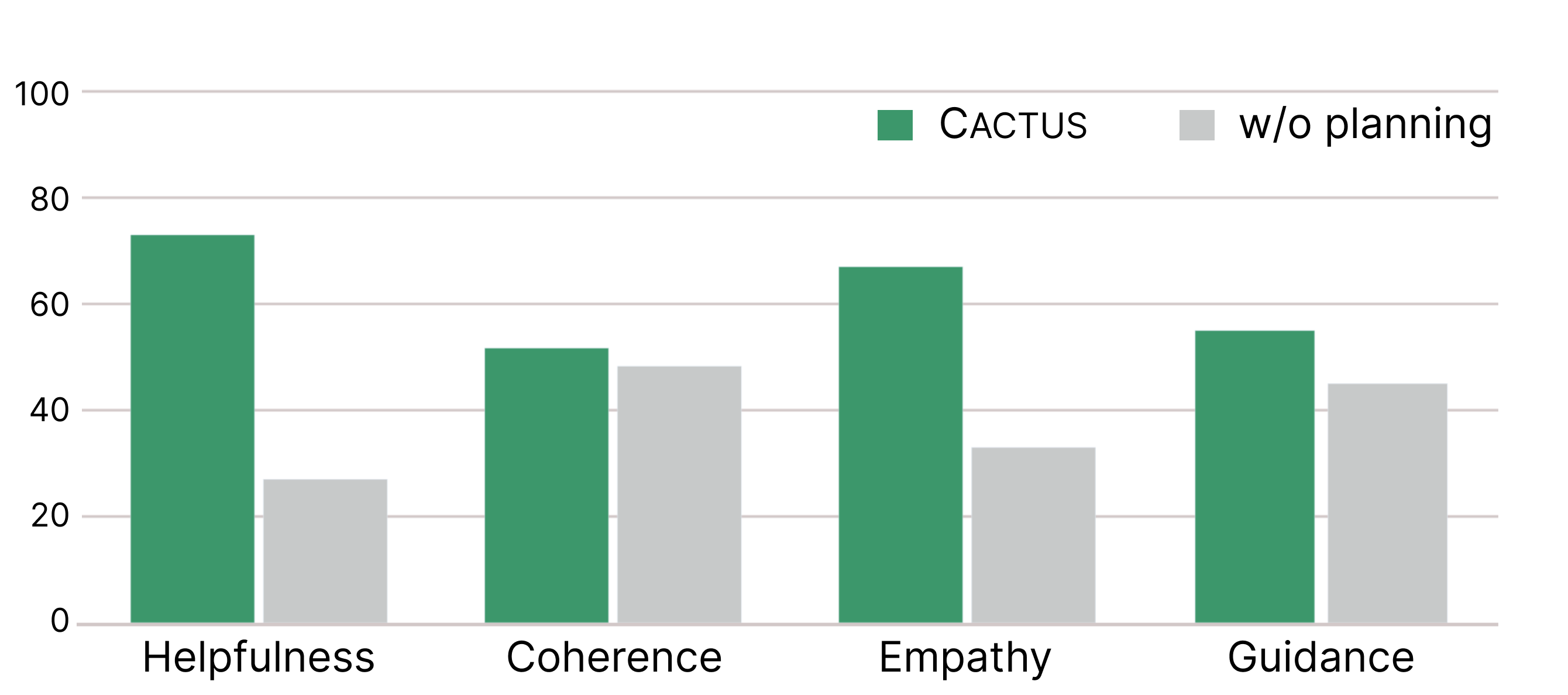}
    \vspace{-10pt}
    \caption{Results of head-to-head comparison between dialogues from \ourdatas{} and planning with out CBT techniques. There are statistically significant differences in \textit{Helpfulness} and \textit{Empathy} ($p < 0.05$).}
    \label{fig:data_quality_ablation}
\end{figure}

%% file: latex/4-experiments.tex
\section{Experiments}
The assessment for the quality of counseling needs to consider both the abilities of counselor and the psychological changes in the client. However, the existing methods, such as automatic or single-turn evaluations, are not suitable for evaluating the abilities of counselor~\citep{smith-etal-2022-human, liu2023chatcounselor}. Therefore, we propose an evaluation framework, \textsc{CounselingEval}.

\input{tables/main_exp_2}
\input{tables/ab_exp_1}

\subsection{\textsc{CounselingEval}: A Psychological Counseling Evaluation Framework}
\paragraph{Method.}
We introduce \textsc{CounselingEval}, an evaluation framework designed to assess counseling skills through multi-turn conversation simulations.
Firstly, to model AI clients like real clients, we construct a set of client information, which includes intake form, attitude, and initial utterance and consists of a total of 450 instances.\footnote{Detailed information is provided in Appendix~\ref{app:counseling_eval}.}
Then, following the Cognitive Therapy Rating Scale (CTRS)~\citep{aarons2012adaptation}, we devise a set of criteria to assess both general counseling and CBT-specific skills.
Finally, we modify the Positive and Negative Affect Schedule (PANAS)~\citep{watson1988development} to assess the effectiveness of counseling from the client's perspective, measuring changes in the client's positive/negative emotions before/after counseling sessions.
For evaluation, we use G-Eval~\citep{liu2023gpteval} to assess each criterion in the Likert-scale with a scoring rubric.

\paragraph{Cognitive Therapy Rating Scale.} 
To evaluate the counselor agent's counseling skills, we utilize CTRS, recognized as the gold standard for measuring counseling effectiveness~\citep{aarons2012adaptation}. CTRS includes criteria for evaluating both general counseling and CBT-specific skills. We select three criteria each for assessing general counseling skills and CBT-specific skills, tailored to our context.\footnote{Details regarding the rationale for selecting these criteria are provided in Appendix~\ref{app:ctrs}.} For each criterion, scores ranging from 0 to 6 are assigned.
The metrics to assess \textbf{general counseling skills} are as follows: 
\begin{itemize}
    \item \textbf{Understanding}: How accurately does the therapist demonstrate understanding of the client's issues and concerns?
    \item \textbf{Interpersonal Effectiveness}: How effective is the therapist in maintaining a positive and therapeutic relationship with the client?
    \item \textbf{Collaboration}: To what extent does the therapist engage the client in collaborative goal-setting and decision-making?
\end{itemize}
 Additionally, the metrics to assess \textbf{CBT-specific skills} are as follows:
 \begin{itemize}
    \item \textbf{Guided Discovery}: How effectively does the therapist use guided discovery techniques to facilitate client self-reflection and insight?
    \item \textbf{Focus}: How well does the therapist identify and address the client’s key cognitions or behaviors that need change?
    \item\textbf{Strategy}: How appropriate and coherent is the therapist's strategy for promoting change in the client's problematic behaviors or thoughts?
\end{itemize}

\input{tables/panas_exp}
\input{figures/ablation_method}
\paragraph{Positive and Negative Affect Schedule.}
The effectiveness of counseling can also be assessed using the Positive and Negative Affect Schedule (PANAS), a commonly used tool for evaluating the effectiveness of counseling. PANAS measures changes in a client's emotions before and after counseling sessions~\citep{watson1988development}. Typically, PANAS consists of two components: (1) evaluating \textbf{positive emotions} and (2) evaluating \textbf{negative emotions}. Each of these components contains 10 items, and participants rate the extent to which they have experienced each emotion on a scale of 1 to 5.\footnote{Details regarding the specific emotions that compose the positive and negative emotions are provided in Appendix~\ref{app:panas}.} Total scores for positive and negative emotions are calculated based on the sum of ratings, respectively. Counseling is deemed more effective if there is an increase in positive emotion scores and a decrease in negative emotion scores after the counseling sessions. Given LLMs' capability to comprehend others' mental states~\citep{kosinski2024evaluating}, we leverage intake forms to infer the client's emotional state before counseling and predict changes in their emotional state after receiving counseling.

\subsection{Experiments Setup}
In \textsc{CounselingEval}, as GPT-4 incurs high API costs, we utilize GPT-3.5-Turbo for the AI clients. However, for the G-Eval of multi-turn conversations, we employ GPT-4o to ensure the reliability of the evaluation.

\paragraph{Counselor agents.}
For comparing the utility of \textsc{Cactus} compared to other datasets, we train LLaMA-2-chat-7B~\citep{touvron2023llama} and LLaMA-3-Instruct-8B~\citep{meta2024llama3} on Psych8k~\citep{liu2023chatcounselor} and SmileChat~\citep{touvron2023llama}. Additionally, we conducted experiments comparing the performance of the model trained on \textsc{Cactus} with CBT-LLM, a model presented in \citep{na2024cbt}. CBT-LLM only provides the model trained on Baichuan-7B without disclosing the dataset. Therefore, unlike other works, it is difficult to make a fair comparison by training the same model on a provided dataset. Instead, we directly used the model provided by CBT-LLM for inference. We use \textsc{Cactus} to train \textsc{Camel} (\textbf{\underline{C}}BT-\textbf{\underline{a}}ug\textbf{\underline{m}}ented couns\textbf{\underline{e}}ling mode\textbf{\underline{l}}) that can perform the following tasks: (1) Selecting the appropriate CBT technique and generating a counseling plan based on the information of client and reason for counseling; (2) Conducting psychological counseling sessions with the client according to the generated counseling plan. Details for training and inference are in the Appendix~\ref{app:experiments}.

\subsection{Results}
\paragraph{Main result.}
The results of the general counseling metrics and counseling technique metrics are presented in Table~\ref{tab:main_exp}. We observe that training with \textsc{Cactus} enhances both general counseling skills and the quality of counseling techniques. Despite SmileChat being trained on a larger dataset compared to Psych8k, Psych8k demonstrates better performance. This superior performance can be attributed to the higher quality of Psych8k, as it is based on real counseling conversations. Moreover, model training with \textsc{Cactus} shows better results in both general counseling skills and CBT-specific skills compared to the CBT-LLM, which is trained on a dataset incorporating CBT. Therefore, the results of our models indicate that \textsc{Cactus} is a high-quality dataset, closely resembling real counseling scenarios.

\paragraph{Effectiveness of planning with CBT.}
To understand the effect of planning with CBT techniques on counseling, we conduct experiments using various methods with GPT-3.5-Turbo. As shown in Table~\ref{tab:ablabtion_planning}, we observe that planning without CBT performs slightly better than not planning at all. However, planning with CBT techniques demonstrates a significant improvement over other methods. Especially, a big gap of performance in \textit{Strategy} means that counselor agents are proficient in using evidence-based strategies to foster meaningful and lasting changes in the client's thought patterns and behaviors.
 
\paragraph{Evaluation from the client’s perspective.}
Evaluating the effectiveness of counseling solely based on utterances may be insufficient. Therefore, we use PANAS to measure counseling efficacy by assessing changes in client emotions. Table~\ref{tab:panas_exp} indicates that \textsc{Camel} shows strong effectiveness in enhancing positive emotions but has limitations in reducing negative emotions. This could be attributed to our approach of guiding clients to explore thought patterns from various perspectives rather than directly changing them.

\paragraph{Validity of utilizing PANAS for evaluation.}
To the best of our knowledge, conducting PANAS with an AI client to measure the effectiveness of counseling is an approach pioneered by our work. As shown in Table~\ref{tab:panas_exp}, we observe a significant magnitude of emotional change for clients with a positive attitude compared to those with a negative attitude. Moreover, the average positive and negative scores obtained from CAMEL-LLAMA3, 28.4 and 22.5 respectively, align closely with results from studies involving human subjects, which recorded mean scores of 33.3 (SD ± 7.2) for positive emotions and 17.4 (SD ± 6.2) for negative emotions~\citep{watson1988development}. These similarities underscore the validity of employing PANAS to assess AI clients, suggesting it is a reasonable approach.

\paragraph{Effectiveness of filtering.}
To ensure the quality of \textsc{Cactus}, we go through the filtering process. We conduct an ablation study to evaluate the effectiveness of the filtering process. Figure~\ref{fig:ablation_method} shows that the filtering process significantly improves counseling skills. When augmenting datasets through various counseling techniques, there is a possibility of choosing improper techniques. This result demonstrates the importance of eliminating such erroneous technique selections.

\subsection{Expert Evaluation}
We conduct an expert evaluation with our domain expert co-authors to validate the effectiveness of \textsc{Camel} which is demonstrated by G-Eval and the appropriateness of using LLMs for evaluating psychological counseling skills. We randomly select 50 dialogues each from Psych-\textsc{LLaMA3} and \textsc{Camel-LLaMA3}, resulting in a total of 100 evaluation instances. In consultation with experts, we decide to evaluate the overall scores from both the general counseling and the CBT-specific skills.

\input{tables/correlation}
\paragraph{Can GPT-4o evaluate the effectiveness of counseling?}
To validate the trustworthiness of LLM-based evaluations, we compare the correlation between the expert and the other evaluators (GPT-4o and non-expert). For non-expert evaluation, we conduct a human evaluation on AMT. In Table~\ref{tab:correlation}, GPT-4o exhibits a higher correlation coefficient than non-expert evaluators. This indicates that LLM-based evaluations have the potential to effectively assess the effectiveness of counseling and can serve as an alternative method to replace expert evaluations, which often face challenges such as scalability and resource constraints.

\paragraph{Results of qualitative evaluation.}
After the expert evaluations, psychologists note that there have been quality differences, however, there are aspects that do not meet psychological standards, resulting in similar scores. Therefore, they suggest that a qualitative evaluation is more appropriate for comparison than a quantitative evaluation. The results of the quantitative evaluation are in Appendix~\ref{app:quantitative_eval}. 
For qualitative evaluation, three psychologists compare the counseling conversation between Psych8k-\textsc{LLaMA3} and \textsc{Camel-LLaMA3}. As a result, Psych8k-\textsc{LLaMA3} gives the impression of preaching rather than collaboratively exploring and finding a solution with the client. In other words, it only suggests a method without engaging in interactive counseling. 
In contrast, \textsc{Camel-LLaMA3} shows superior counseling skills compared to other models by exploring the issues of the client and showing empathy for the client's emotions. However, it exhibits limitations in its tendency to ask superficial questions and a lack of personalized exploration.

\subsection{Future Direction of Counselor Agent}
We aim to present our insights based on discussions with our psychological expert co-authors:
\paragraph{\textbf{Counselors should not overly empathize with the client.}}
Counselors aim to understand their clients' situations more accurately by facilitating better expression through empathy. Therefore, instead of being overly empathetic and anticipating beyond what the client expresses, counselors should empathize specifically with what the client discusses after hearing the details of the problem.
\paragraph{\textbf{Counselors are questioners, not answer-providers.}}
Rather than delivering one-sided solutions, counselors should guide clients to self-realization and find appropriate solutions themselves. This involves guiding clients through reflective questioning to help organize their thoughts.

Previous research has focused on developing counselors to give more empathetic and good answers. We would recommend that future counseling research shift its perspective to view counselors as questioners, moving away from the traditional role of counselors as answer providers.

%% file: tables/main_exp_2.tex
\begin{table*}[t!]
\centering
\resizebox{0.9\textwidth}{!}
{
\begin{tabular}{lcccccccc}
    \toprule
    \multirow{2.5}{*}{Model}           & \multicolumn{3}{c}{General Counseling Skills} & \multicolumn{3}{c}{CBT-specific Skills}  \\ \cmidrule(lr){2-4} \cmidrule(lr){5-7}
    & Understanding & Interpersonal Eff. & Collaboration & Guided Discovery  & Focus & Strategy \\
    \midrule
    CBT-LLM  & 4.00 & 5.12 & 3.79 & 2.66 & 3.95 & 4.02 \\
    \midrule
    Psych8k\textsc{-llama2}  & 3.99 & 4.78 & 3.92 & 2.79 & 3.98 & 4.03 \\
    SmileChat\textsc{-llama2}  & 3.94 & 4.29 & 3.49 & 2.32 & 3.86 & 3.65 \\
    \textbf{\textsc{Camel-llama2}} & \textbf{4.20} & \textbf{5.41} & \textbf{4.42} & \textbf{3.80} & \textbf{4.07} & \textbf{4.81} \\
    \midrule
    Psych8k\textsc{-llama3}  & 3.96 & 4.97 & 3.69 & 2.90 & 3.90 & 4.03 \\
    SmileChat\textsc{-llama3}  & 3.97 & 4.54 & 3.54 & 2.35 & 3.89 & 3.75 \\
    \textbf{\textsc{Camel-llama3}} & \textbf{4.42} & \textbf{5.97} & \textbf{4.81} & \textbf{4.40} & \textbf{4.11} & \textbf{5.11} \\
    \bottomrule
\end{tabular}
}
\caption{Results of \textsc{CounselingEval} on general counseling and CBT-specific skills for the trained models. All models are fine-tuned on counseling dataset and the best results for each base model are \textbf{bolded}.}
\label{tab:main_exp}
\end{table*}

%% file: tables/ab_exp_1.tex
\begin{table*}[!ht]
\centering
\resizebox{0.98\textwidth}{!}
{
\begin{tabular}{llccccccc}
    \toprule
    \multirow{2.5}{*}{Model} & \multirow{2.5}{*}{Method}    & \multicolumn{3}{c}{General Counseling Skills} & \multicolumn{3}{c}{CBT-specific Skills}  \\ 
    \cmidrule(lr){3-5} \cmidrule(lr){6-8}
    &  & Understanding & Interpersonal Eff. & Collaboration & Guided Discovery  & Focus & Strategy \\
    \midrule
     \multirow{3}{*}{GPT-3.5-Turbo} & w/o planning & 4.02 & 5.47 & 4.01 & 3.29 & 3.99 & 4.07 \\
     & planning w/o CBT  & 4.00 & 5.51 & 4.02 & 3.31 & 4.00 & 4.07 \\
     & planning w/ CBT & \textbf{4.03} & \textbf{5.63} & \textbf{4.10} & \textbf{3.44} & \textbf{4.17} & \textbf{4.62} \\
    \bottomrule
\end{tabular}
}
\caption{Results of \textsc{CounselingEval} on general counseling and CBT-specific skills for GPT-3.5-Turbo with different methods. The best results for each base model are \textbf{bolded}.}
\label{tab:ablabtion_planning}
\vspace{-10pt}
\end{table*}

%% file: tables/panas_exp.tex
\begin{table*}[t!]
\centering
\small
\resizebox{0.82\textwidth}{!}
{
\begin{tabular}{lcccccc}
    \toprule
    \multirow{2.5}{*}{Model}           & \multicolumn{2}{c}{Positive Attitude} & \multicolumn{2}{c}{Neutral Attitude} & \multicolumn{2}{c}{Negative Attitude}\\ \cmidrule(lr){2-3} \cmidrule(lr){4-5} \cmidrule(lr){6-7}
                                    & positive $\uparrow$ & negative $\downarrow$ & positive $\uparrow$ & negative $\downarrow$   & positive $\uparrow$ & negative $\downarrow$ \\
                                    \midrule
    \text{GPT-3.5-Turbo} & +0.71 & \textbf{-0.55} & +0.73 & -0.40 & +0.39 & \textbf{-0.48} \\
    \text{Psych8k-\textsc{llama3}} & +0.74 & -0.34 & +0.58 & -0.30 & +0.54 & -0.43 \\
    \text{\textsc{Camel-llama3}} & \textbf{+1.17} & -0.49 & \textbf{+0.94} & \textbf{-0.41} & \textbf{+0.65} & -0.10 \\
    \bottomrule
\end{tabular}

}
\caption{Results of \textsc{CounselingEval} on PANAS across various models. This table illustrates the shifts in positive and negative emotions before and after counseling sessions, categorized by the client’s attitude.}
\vspace{-10pt}
\label{tab:panas_exp}
\end{table*}

%% file: figures/ablation_method.tex
\begin{figure}[t!]
    \centering
    \includegraphics[width=0.94\linewidth]{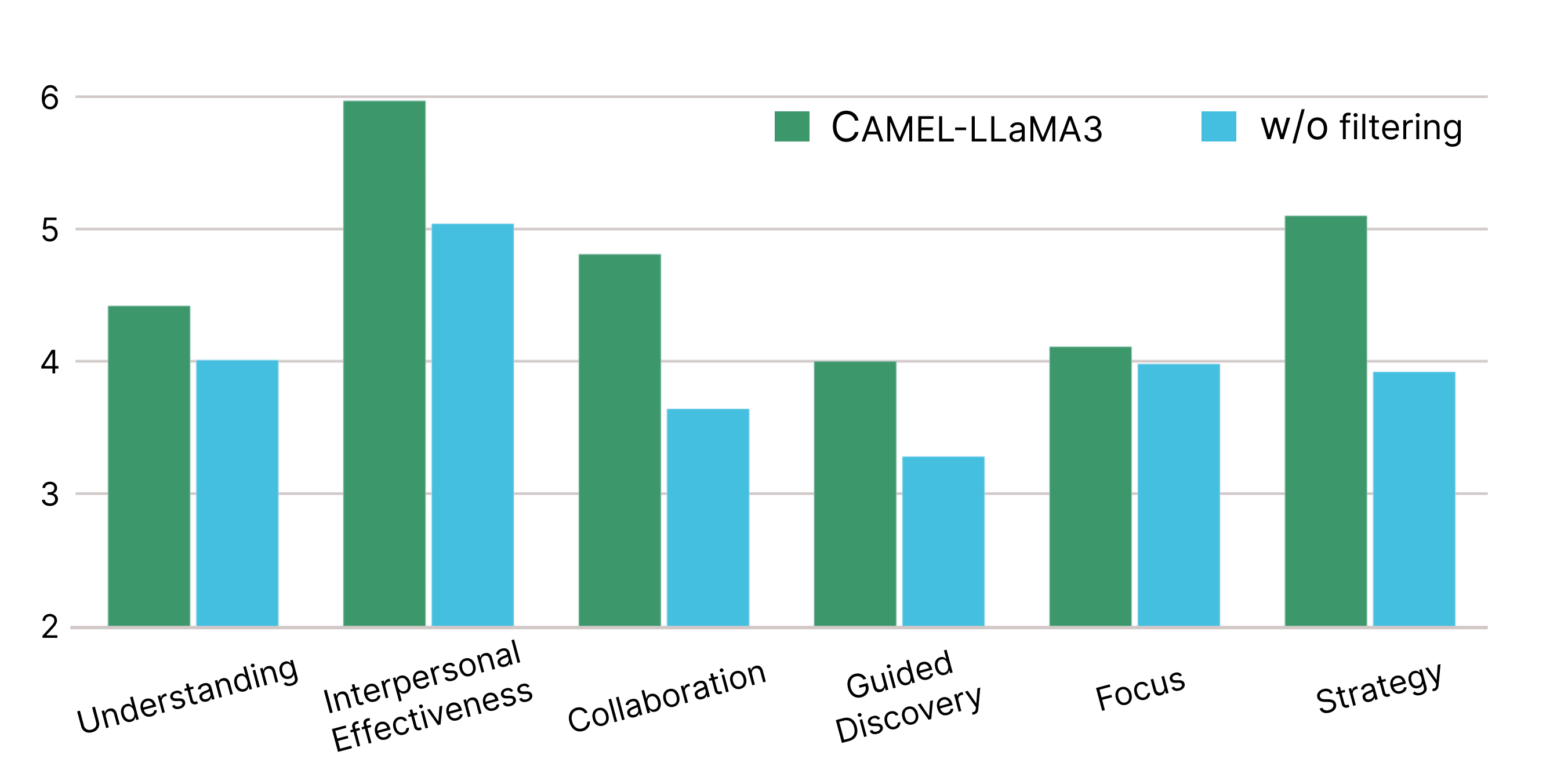}
    \caption{Results of ablation study on general counseling and CBT-specific skills for filtering.}
    \label{fig:ablation_method}
    \vspace{-10pt}
\end{figure}

%% file: tables/correlation.tex
\begin{table}[!t]
\centering
\resizebox{\linewidth}{!}
{
\begin{tabular}{lcccccc}
    \toprule
    \multirow{2.5}{*}{Evaluator} & \multicolumn{3}{c}{General Counseling}  & \multicolumn{3}{c}{CBT-specific} \\
    \cmidrule(lr){2-4} \cmidrule(lr){5-7}
    & $r$ & $\rho$ & $\tau$ & $r$ & $\rho$ & $\tau$ \\
    \midrule
    Non-expert & 0.51 & 0.17 & 0.15 & 0.09 & 0.08 & 0.07\\
    GPT-4o & \textbf{0.60} & \textbf{0.19} & \textbf{0.16} & \textbf{0.65} & \textbf{0.65} & \textbf{0.61} \\
    \bottomrule
\end{tabular}
}
\caption{The correlations between the expert and other evaluators are represented by 
$r$, $\rho$, and $\tau$, which indicate the Pearson, Spearman, and Kendall's Tau correlation coefficients, respectively.}
\vspace{-10pt}
\label{tab:correlation}
\end{table}

%% file: latex/5-related-work.tex
\section{Related Work}

\subsection{Cognitive Behavioral Therapy}
People with depression or anxiety form negative, irrational thoughts that reinforce negative beliefs about themselves, others, and the world~\citep{beck2020cognitive}. CBT aims to break this cycle by identifying and challenging these automatic thoughts and core beliefs~\citep{LONGMORE2007173}.
In CBT, counselors first help clients recognize unhelpful thoughts. Then they guide clients to challenge and correct these distortions using CBT techniques, gradually reconstructing more positive automatic thoughts and beliefs~\citep{fenn2013key}.
Our dataset simulates CBT dialogue interactions focusing on clients with depression and anxiety disorders treated through CBT~\citep{carroll2017cognitive}.

\subsection{Psychological Counseling}
While there is growing interest in using LLMs for counseling, maintaining therapeutic consistency over multiple dialogue turns remains challenging. Existing work largely focuses on single-turn counseling strategies~\citep{sharma2023cognitive, maddela2023training, sun-etal-2021-psyqa}. Although some attempts have been made at multi-turn counseling, they have limitations-achieving only 2-3 turns~\citep{xiao2024healme} or simply extending single-turn interactions without capturing authentic multi-turn client interactions~\citep{qiu2023smile}.
To address this gap, our work presents a counseling dialogue dataset that applies CBT techniques across multiple turns while maintaining realism and closeness to real-world counseling scenarios. This aims to enable more natural, consistent therapeutic dialogues with LLMs over an extended interaction.

%% file: latex/6-conclusion.tex
\section{Conclusions}
We introduce \textsc{Cactus}, a large-scale synthetic dataset of counseling dialogue. It aims to provide realistic multi-turn conversations by having the AI counselor and client exhibit real-life behaviors through simulation before sessions. Including diverse client personas with varying counseling attitudes and generating conversations based on different counseling strategies counselors may use, results in a diverse dataset. \textsc{Camel}, trained on \textsc{Cactus} demonstrated high performance across all domains. Additionally, we propose \textsc{CounselingEval}, an evaluation framework that simulates dialogues between an AI counselor and AI clients modeling real clients. It applies established psychological criteria like CTRS and PANAS to these simulated dialogues, enabling human-aligned evaluation of counseling conversations.

\section*{Limitations}
In actual counseling sessions, it is common for sessions to last around an hour each, with a total of about 10 sessions typically conducted. While our setup involves relatively long multi-turn interactions compared to other datasets, it's still considerably shorter than real-life counseling sessions. This is a limitation, as we haven't yet considered multi-session interactions, which are integral to real counseling practices.

While it is true that in some cases, appropriate CBT techniques can be applied based on the intake form filled out by the client, in reality, counselors dynamically choose or modify strategies based on the responses of the client during the conversation. However, we adhere strictly to the initial selection of CBT techniques and planning methods, which deviates slightly from real counseling practices.

In future work, it would be beneficial to incorporate longer conversations and consider multi-session interactions to make our approach more akin to real-world scenarios. Additionally, it is recommended to adopt a more flexible approach in selecting counseling strategies based on client responses.

\section*{Ethical Considerations}
Interventions in mental health demand careful ethical examination from the standpoints of safety, privacy, and bias mitigation.

\paragraph{Safety.}
 There is a possibility that, despite its helpful intentions, AI could have negative impacts on individuals with mental health challenges. While our model has shown some degree of therapeutic effectiveness, we believe it should be used under the supervision of a professional rather than being employed solely in counseling sessions. Our primary target is individuals experiencing mild depression and anxiety; therefore, we advise to avoid using ours for those with more severe psychological issues beyond its intended scope.
While our AI model has demonstrated therapeutic potential, there is a possibility that it could unintentionally cause harm to individuals grappling with mental health challenges. Consequently, we strongly advocate for the use of our model under professional supervision, rather than as a standalone counseling tool. Our model's intended scope encompasses individuals experiencing mild depression and anxiety; therefore, we advise against its unsupervised use for those suffering from more severe psychological conditions that extend beyond its capabilities.

\paragraph{Privacy.}
To preserve privacy and maintain ethical integrity, we deliberately avoid utilizing real client data in the process of simulation counseling scenarios. Instead, we employ publicly available datasets purposefully curated for research endeavors. These datasets are constructed through crowdsourcing information from psychological experts, not by collecting data from actual clients. This approach mitigates ethical concerns surrounding personal identification and confidentiality breaches. Furthermore, the information provided by the psychological experts is generalized and does not reflect any specific individual's psychological profile, thereby upholding the ethical standards for data usage in mental health research.

\paragraph{Bias.}
Despite our efforts to create a diverse and representative dataset by considering factors such as gender, age, and occupation, the potential for demographic bias persists. The dataset creation process involve the use of ChatGPT, which itself is trained on vast amounts of internet data that may contain inherent societal biases and prejudices reflected in online content. Moreover, although we assigned names randomly from a comprehensive directory to minimize the risk of identifying individuals, there is an ever-present need for vigilance to avoid unintentional biases arising from these selections. Our model, trained on this dataset, could inadvertently acquire and propagate these biases, potentially resulting in the over- or under-representation of certain demographic groups. Consequently, it is imperative to approach the deployment of our model with utmost caution. Continuous monitoring and proactive adjustments are necessary to identify and rectify any emergent biases. Ethical deployment also necessitates transparent communication with users regarding the potential limitations and biases of the AI system.

\section{Acknowledgments}
This work was supported by Institute of Information \& Communications Technology Planning \& Evaluation (IITP) grant funded by the Korean government (MSIT)(No.RS-2020-II201361, Artificial Intelligence Graduate School Program (Yonsei University)), (2022-0-00077, RS-2022-II220077, AI Technology Development for Commonsense Extraction, Reasoning, and Inference from Heterogeneous Data), and the ITRC(Information Technology Research Center) support program(IITP-2024-RS-2024-00437102). Jinyoung Yeo is the corresponding author.

%% file: latex/9-appendix.tex
\appendix
\newpage
\section{Cognitive Behavior Therapy (CBT) Technique}
\label{app/cbt_tech}
\subsection{The Types of Patterns}
The types of patterns and the examples of each pattern type are presented in Table~\ref{table:cognitive_distortions_example}. We reference the definition of patterns from \citet{sharma2023cognitive}, and examples for each pattern were sourced from \textsc{PatternReframe}~\citep{maddela2023training}.
\input{tables/cognitive_distortions}

\subsection{CBT Strategies}
\subsubsection{CBT Technique Selection}
According to \citet{beck2020cognitive}, there are 20 types of Cognitive Behavioral Therapy (CBT) techniques. We define and utilize 12 of these techniques as CBT techniques. The selection process is conducted in collaboration with the psychological experts.
Below are the reasons we did not select the 8 CBT techniques.
\begin{itemize}
    \item \textbf{Guided discovery, Socratic questioning}: These techniques are excluded as they represent broader concepts in CBT and are widely used across multiple techniques. 
    \item \textbf{Scaling Questions, Thought experiment}: As these techniques are less frequently utilized in real-world counseling practices, we do not include them in our dataset construction process.
    \item  \textbf{Activity scheduling, Role-playing and Simulation, Practice of Assertive Conversation Skills, Safety behaviors elimination}: We exclude these techniques as they are less suitable for application in remote counseling sessions.
\end{itemize}

\subsubsection{Types of CBT Technique}
The descriptions for the twelve selected CBT techniques can be found in Table~\ref{table:cbt_technique_example}.
\input{tables/cbt_technique}

\section{Details of Experiment on the Challenges of using LLMs}
\label{app:empirical_investigations}
\subsection{Limitations in selecting CBT Technique}
\label{app:desing_consider1}
To compare the CBT techniques selected by humans and LLMs, we conducted an experiment. We provided 150 client thoughts and patterns to three psychological experts and asked them to choose up to three appropriate CBT techniques. Only the techniques chosen by at least two experts were selected as the gold label, with an average of 1.6 techniques per thought-pattern pair. We then presented the same thoughts and patterns, along with the list of CBT techniques, to the LLMs, instructing them to select two suitable techniques for each thought-pattern pair. The results are shown in Figure~\ref{fig:cbt_tech_distribution_chatgpt}. While GPT-4o exhibited less bias compared to GPT-3.5, it still demonstrated a significant level of bias.
\input{figures/chatgpt_cbt}

\subsection{Limitations of LLM as an AI Client and an AI Counselor}
\label{app:desing_consider2}
We conduct an empirical investigation to examine the behaviors exhibited when using LLMs to simulate counselor-client interactions. For this, we utilize the prompts suggested by \citet{na2024cbt} in their study on creating a CBT counseling dataset. The prompts we used are shown in Figure~\ref{fig:app_cbt_llm}.
\input{figures/prompt_cbt_llm}
 
\section{Details on Human Evaluation}
\label{app:data_quality_human}
\subsection{Implementations of Human Evaluation}
To compare \textsc{Cactus} and existing counseling dialogue datasets, we conduct human evaluation via Amazon Mechanical Turk (AMT). Figure~\ref{fig:amt_interface} shows the interface employed for comparative evaluations (\textit{Win/lose}) between two datasets. Detailed instructions and rubrics for each score are included to ensure precise evaluation. For each evaluation, we asked three human annotators to assess 100 samples based on four specified criteria. We compensated each annotator \$0.30 per evaluated sample.

\subsection{Human Evaluation Criteria}
We ask the judges to compare the dialogues based on the following criteria:
\begin{itemize}
    \item{\textbf{Helpfulness} measures the suitability of interpretations and suggestions from a psychological counseling perspective. }
    \item {\textbf{Coherence} measures the logical flow and structure of the session.}
    \item {\textbf{Empathy} measures the ability of the counselor to understand and respond to feelings of the client.}
    \item {\textbf{Guidance} measures the specificity and practicality of suggestions of the counselor.}
\end{itemize}

\section{Details of \textsc{CounselingEval}}
\subsection{Method}
\label{app:counseling_eval}
\textsc{CounselingEval} is a psychological counseling evaluation framework designed to assess the counseling skills of counselor agents through interaction with the AI client.\footnote{In this work, we use \texttt{gpt-3.5-turbo-0125} for the AI client.} 
The test dataset includes detailed client information for the AI client, client information that is accessible to the counselor agent, and an initial utterance to start the counseling session. 
This set comprises 150 distinct client profiles, each presenting three different counseling attitudes (positive, neutral, negative), resulting in a total of 450 instances.
First, the counselor agent generates the next utterance based on the client's information and the initial utterance. Then, the AI agent and the counselor agent proceed with the interactive counseling session. In the case of the planning with the CBT model (\eg, \textsc{Camel}), the process of CBT technique selection and planning is added before generating the first utterance from the counselor.
The counseling session concludes when the AI client generates the end token (\texttt{[END]}). The quality of the generated multi-turn counseling conversation is then evaluated using CTRS and PANAS.

\subsection{Cognitive Therapy Rating Scale (CTRS)}
\label{app:ctrs}
The CTRS evaluates both general counseling skills and CBT-specific skills. Originally, the CTRS consists of six criteria for general counseling skills (agenda, feedback, understanding, interpersonal effectiveness, collaboration, pacing, and efficient use of time) and six criteria for CBT-specific skills (guided discovery, focusing on key cognitions or behaviors, strategy for change, application of cognitive-behavioral techniques, homework). We select three criteria from general counseling skills, which are understanding, interpersonal effectiveness, collaboration, and three criteria from CBT-specific skills, which are guided discovery, focusing on key cognitions or behaviors, strategy for change.

Given that the counseling is conducted through the text, we exclude criteria such as feedback, pacing and efficient use of time (evaluating the ability to conduct counseling within a set timeframe), and homework (assessing the ability to assign tasks for actual behavioral change) as such criteria pertain to nonverbal elements and practical tasks not applicable to text-based counseling. Furthermore, since our counseling sessions commence with a provided intake form, there is no need to evaluate the agenda setting process separately, so we omit the agenda criterion. Additionally, considering that GPT lacks the ability to apply cognitive-behavioral techniques, we exclude this criterion from our evaluation criteria. The prompt used for CTRS scoring is shown in Figure~\ref{fig:app_prompt_CTRS}.
\input{figures/prompt_CTRS}

\subsection{Positive and Negative Affect Schedule (PANAS)}
\label{app:panas}
The Positive and Negative Affect Schedule (PANAS) is a standardized tool that assesses both positive and negative effects that individuals experience either currently or over a specific period. Thus, PANAS can be utilized as a measure to assess counseling by measuring changes in emotions that clients perceive before and after counseling.

Typically, PANAS consists of two parts, each containing 10 items representing positive and negative emotions. Positive emotions include Interested, Excited, Strong, Enthusiastic, Proud, Alert, Inspired, Determined, Attentive, and Active, while negative emotions include Distressed, Upset, Guilty, Scared, Hostile, Irritable, Ashamed, Nervous, Jittery, and Afraid. Participants rate the extent to which they have experienced each emotion on a scale of 1 to 5. The prompt used for PANAS scoring can be found in Figure~\ref{fig:app_prompt_PANAS}.
\input{figures/prompt_PANAS}

\section{Details of \textsc{Cactus}}
\label{app:details_of_cactus}
We provide sample dialogues in Table \ref{tab:cactus_example}, \ref{tab:cactus_example_neutral}, and \ref{tab:cactus_example_negative}. Also, sample thought, patterns, intake form, selected CBT technique, and plan for counseling are presented in Table~\ref{tab:cactus_example_steps}.
\subsection{Client Simulation}
\label{app/client_simulation}
\paragraph{Intake form.}
To simulate realistic clients, we automatically fill out the intake form, which includes questions typically asked by psychological experts during intake sessions. Specifically, we instruct GPT-3.5 to depict the situation of the client with the characteristics seeking therapy. The intake form includes information such as presenting problems, past history, etc. Further details about the intake form can be found in Figure~\ref{fig:app_prompt_intake_form}.

\paragraph{Diverse attitude of client.}
We categorize the attitudes of clients into three types: positive, neutral, and negative. Subsequently, we provide behaviors associated with each attitude to help the model simulate the client with the given attitude more concretely following ~\citet{li2023understanding}. For clients with a neutral attitude, we provide a mix of both positive and negative characteristics. Detailed descriptions of the positive and negative attitudes are provided in Table~\ref{tab:attitude_description}.

\input{tables/attitude_description}

\subsection{Counselor Simulation}
\label{app/counselor_simulation}
As CBT is known as a goal-oriented and structured approach, we add the planning process before the counseling session. Firstly, given thought, patterns and reframed thought, GPT-3.5 model is instructed to choose top-3 CBT techniques that could be used to frame the given thought into reframed thought. Then, GPT-3.5 generates a plan for the counseling session based on the client's intake form and selected CBT techniques. The prompt used for CBT technique selecting and the prompt used for counseling planning are shown in Figure~\ref{fig:app_prompt_cbt} and Figure~\ref{fig:app_prompt_plan} respectively.

\subsection{Dialogue Generation}
\label{app/dialogue_generation}
\paragraph{Two-agent mode vs. Script mode.}
We conduct experiments to compare two-agent mode~\citep{zhou2023sotopia} and script mode~\citep{kim-etal-2023-soda}, aiming to identify which method generates more dialogue. To assess the quality of dataset, we conduct human evaluations on AMT, applying the same criteria listed in Appendix~\ref{app:data_quality_human}. Additionally, we include \textit{Naturalness} as criteria, which evaluates how naturally and smoothly the conversation flows like human-like interactions. As shown in Figure~\ref{fig:two_agent_script}, the script mode outperforms the two-agent mode across most criteria and demonstrates significantly better naturalness. As a result, we generate counseling dialogue with script mode.

\subsection{Filtering}
\paragraph{Data filtering of PatternReframe.}
\label{app:pattern_filtering}
We conduct a filtering process on the PatternReframe to create high-quality data. Focusing on clients with anxiety or depression, we aim to generate counseling dialogues using CBT techniques. Severe mental disorders like delusions or auditory hallucinations are inappropriate for CBT~\citep{beck2020cognitive}. We gathered annotations from psychological experts on negative thoughts in the PatternReframe and used these annotations as few-shot exemplars to the \texttt{gpt-3.5-turbo-0125} model, improving its ability to classify thoughts and apply appropriate CBT techniques.

\paragraph{Basic filtering.}
With generated dialogues, we first filter out dialogues based on two criteria: (1) the number of turns and (2) the format of dialogue. As psychological counseling with CBT requires more turns than chit-chat, we set the appropriate number of turns as 20-35 turns and remove dialogues that are too short or long. Moreover, we discard dialogues without speaker prefixes by using lexical pattern matching. After basic filtering, 96.36\% of the initial dialogues remain, which are 35,252 dialogues.

\paragraph{Filtering with CTRS score.}
\label{app:filtering}
The Cognitive Therapy Rating Scale (CTRS) is an observer-rated measure that is utilized to assess how well a counselor performs cognitive therapy~\citep{beck2020cognitive}. 
From the 11 CTRS items, we select the 6 most appropriate criteria for our dataset creation and evaluate the generated counseling dialogues based on these criteria. By filtering with CTRS score, we aim to construct a high-quality psychological counseling dataset. We rate the dialogues on a scale of 0 to 6 using GPT-3.5 model as our judge. We average the scores of six criteria and filter out dialogues when the average score is smaller than 5.0. Finally, 86.31\% of the dialogues remain, which form the 31,577 dialogues in \ourdatas{}. We provide a sample dialogue in Table~\ref{tab:cactus_example}.

\section{Details of Experiments}
\label{app:experiments}
\input{figures/prompt_camel_cbt}
\input{figures/prompt_camel}
\subsection{Training}
\label{app:training}
To ensure a fair comparison among the datasets and mitigate any discrepancies arising from model selection, we do not use the models provided by \citet{liu2023chatcounselor} and \citet{qiu2023smile}. 
For training, we employ QLoRA \citep{dettmers2023qlora} to fine-tune our model using 4-bit quantization. We set the dimension of low-rank matrices to 64 and alpha to 16. The DeepSpeed library\footnote{https://www.deepspeed.ai} facilitates the training with a learning rate of 2e-4. The model is trained for 5 epochs on the Psych8k dataset and for 2 epochs on the \textsc{SmileChat} and \textsc{Cactus} datasets. For training \textsc{Camel}, we use the templates in Figure~\ref{fig:app_prompt_camel_cbt} for planning with CBT techniques and Figure~\ref{fig:app_prompt_camel} for generating utterance.

\subsection{Inference}
For evaluating the dialogues using LLM (\ie, G-Eval), we use GPT-4o and adopt temperature sampling with  $T = 0.0$.
Additionally, for generating responses of an AI client and a counselor, we adopt temperature sampling with $T = 0.7$.
To achieve higher throughput during inference, we leverage the vLLM library.\footnote{https://docs.vllm.ai}

\subsection{Terms and License.} 
For our implementation and evaluation, we use Huggingface library\footnote{\url{https://huggingface.co/}} and vLLM library. 
Both libraries are licensed under Apache License, Version 2.0.
We have confirmed that all of the artifacts used in this paper are available for non-commercial scientific use.

\section{In-depth Analysis}
\subsection{Analysis of \textsc{Cactus}}
\label{app:analysis_cactus}
\paragraph{Diversity of \textsc{Cactus}.}
We aim to create a dataset encompassing diverse client cases, capturing a range of issues, thought patterns, and attitudes toward counseling. Table~\ref{tab:cactus_statistics} shows that the dataset covers a wide range of issues that clients face, from relationships (20.83\%) and career concerns (17.50\%) to anxiety (14.17\%) and hobbies (10.83\%). This variety ensures that the dataset encompasses many aspects of clients' lives. Clients from various age groups are represented, from teenagers (10-19 years, 1.83\%) to seniors (80-89 years, 0.32\%). The spread across different age brackets (most notably 20-29 years at 22.18\% and 30-39 years at 24.24\%) indicates a wide age range, enhancing the dataset's diversity in terms of life stages and generational perspectives. Further details can be found in Table~\ref{tab:cactus_statistics}.

\subsection{Main Results}
\input{figures/app_smile_cactus}
\paragraph{SmileChat vs. \textsc{Cactus}.}
Since SmileChat is a Chinese dataset, directly comparing it with models trained on the English dataset \textsc{Cactus} might be unfair. Therefore, we translate the results of \textsc{Camel-LLaMA3} on \textsc{CounselingEval} into Chinese and compare them with the CounselingEval results of SmileChat-\textsc{LLaMA3}. As shown in Figure~\ref{fig:app_smile_cactus}, even though there is a possibility of degradation in the response quality due to the translation, \textsc{Camel-LLaMA3} outperforms SmileChat-\textsc{LLaMA3} in all axes.

\subsection{Expert Evaluation}
We conduct an expert evaluation of 50 sampled counseling conversations for each model, randomly selecting diverse client information and attitudes from our test set with 450 samples.
\paragraph{Quantitative evaluation.}
\label{app:quantitative_eval} 
From the head-to-head comparison of the counseling ability between \textsc{Camel-LLaMA3} and Psych8k-\textsc{LLaMA3}, we found that the win rate of \textsc{Camel-LLaMA3} 50\%. However, psychological experts explain that these results may not accurately reflect good performance. While our model may show better performance than Psych8k-\textsc{LLaMA3}, experts conclude that both models do not fully meet the standards of professional psychological counseling. This suggests that quantitative evaluations alone are insufficient for assessing their effectiveness comprehensively. Furthermore, experts underscore the importance of qualitative evaluations in driving the development of psychological counseling models.

\section{Case Study}
\label{app:case_study}
\subsection{Problems of Using ChatGPT as a Counselor and a Client Agent}
Figure~\ref{fig:app_problem} shows the examples of the issues identified for design considerations of \textsc{Cactus}, as discussed in Section~\ref{sec:design}.

\subsection{Effectiveness of planning with CBT}
To confirm the effectiveness of planning with the CBT technique, which is one of our main ideas, we present results of ablation studies in Figure~\ref{fig:data_quality_ablation} and Figure~\ref{fig:ablation_method} from the previous section. Additionally, the specific case demonstrating the effectiveness of planning with CBT techniques is illustrated in Figure~\ref{fig:app_chatgpt}.

\subsection{Psych8k vs. \textsc{Cactus}}
Psych8k-\textsc{LLaMA3} often presents too much information at once, which can overwhelm clients and lead to ineffective interaction between the counselor and client, with sessions sometimes concluding abruptly. In contrast, \textsc{Camel-LLaMA3} gradually facilitates the counseling process, effectively drawing out the client's issues and fostering a more engaging and supportive environment, as shown in Figure~\ref{fig:app_psych_camel}.

\input{tables/statistics}
\input{tables/dataset_example1}
\input{tables/dataset_example2}
\input{tables/dataset_example3}
\input{tables/dataset_example_steps}

\input{figures/prompt_intake_form}
\input{figures/prompt_cbt}
\input{figures/prompt_planning}
\input{figures/prompt_dialog}
\input{figures/prompt_ai_client}

\input{figures/amt_interface}
\input{figures/app_problem_example}
\input{figures/app_chatgpt_example}
\input{figures/app_psych_camel_example}

%% file: tables/cognitive_distortions.tex
\begin{table*}[t!]
\centering
\small
\begin{tabular}{p{3.2cm} p{10.0cm}}
\toprule
\textbf{Patterns} & \textbf{Definition and Example of Unhelpful Thought with Cognitive distortion}  \\ 
\midrule
\multirow{2}{*}{Catastrophizing} & Focusing on the worst-case scenario \\
& \textit{My mom hasn't come home from work yet. I hope the store isn`t getting robbed!} \\ 
\midrule
\multirow{2}{*}{Discounting the positive} & When something good happens, you ignore it or think it doesn’t count \\
& \textit{My restaurant is the most popular in my city, but that's just luck.} \\
\midrule 
\multirow{2}{*}{Labeling} & Defining a person based on one action or characteristic \\
& \textit{I fell off my skateboard yesterday, I'm a terrible athlete.} \\
\midrule
\multirow{2}{*}{Mental Filtering} & Getting “stuck” on a distressing thought, emotion, or belief \\
& \textit{It's nice to enjoy the sea breeze when you live near the ocean but it's not worth it when you think of all the sand getting dragged into your home and all the tourists making so much noise at the beach.} \\
\midrule
\multirow{2}{*}{Mind Reading} & Make assumptions about the thoughts, feelings, or intentions of others based on one’s perceptions or interpretations \\
& \textit{I auditioned for the surf team and the coach avoided me. I am sure it is because he does not like my skills.} \\
\midrule
\multirow{2}{*}{Fortune Telling}  & Trying to predict the future. Focusing on
one possibility and ignoring other, more likely outcomes \\
& \textit{I didn't make it to Yellowstone this year, I am never going to go to that park.} \\
\midrule
\multirow{2}{*}{Personalization} & Taking things personally or making them about you \\
& \textit{My sister was not happy with the makeup look I did for her. I am a bad artist.} \\
\midrule
\multirow{2}{*}{All-or-nothing thinking} &  Thinking in extremes  \\
& \textit{The school Christmas choir concert got canceled. This holiday season is ruined.} \\
\midrule 
\multirow{2}{*}{Overgeneralization} & Jumping to conclusions based on one experience. \\
& \textit{My nephews didn't want to spend the weekend with me this week. I must not be as good of an aunt as I thought.} \\
\midrule
\multirow{2}{*}{Should statements} &  Setting unrealistic expectations for yourself. \\
& \textit{I prefer texting over phone calls. People should never call me and expect me to answer.} \\
\bottomrule
\end{tabular}
\caption{Examples of patterns, negative thoughts, and examples from PatternReframe dataset. Definition is in regular font, while Example is in \textit{italics}.}
\label{table:cognitive_distortions_example}
\end{table*}

%% file: tables/cbt_technique.tex
\begin{table*}[t!]
\centering
\small
\begin{tabular}{p{3.2cm} p{10.0cm}}
\toprule
\textbf{CBT Technique} & \textbf{Description}  \\ 
\midrule
Efficiency Evaluation & Assists individuals in evaluating the usefulness of their thoughts or beliefs, analyzing how practical or detrimental they are in real-life situations. \\
\midrule
Pie Chart Technique & Used for individuals experiencing excessive self-blame or responsibility, visually representing the contribution of various factors to a specific event or outcome. \\
\midrule
Alternative Perspective & Involves asking clients how others might think in similar situations, encouraging consideration of different interpretations. \\
\midrule
Decatastrophizing  
& Aims to reduce the tendency to imagine the worst-case scenario by evaluating the actual likelihood of the feared outcome and preparing for coping strategies. \\ 
\midrule
Pros and Cons Analysis &
Analyzes the advantages and disadvantages of specific thoughts or beliefs, fostering a more balanced evaluation. \\
\midrule
Evidence-Based Questioning & Guides clients to find evidence supporting or contradicting their thoughts, promoting a more evidence-based approach to thinking. \\
\midrule
Reality Testing & Explores how well clients' thoughts align with reality, helping them distinguish between thoughts and actual experiences. \\
\midrule
Continuum Technique & Positions clients' experiences between two extreme situations, encouraging a more nuanced evaluation of situations. \\
\midrule
Changing Rules to Wishes & Replaces strict rules or arbitrary attitudes with realistic hopes or wishes. \\
\midrule
Behavior Experiment & Involves trying out new behaviors in specific situations to challenge and modify negative beliefs. \\

\midrule
Problem-Solving Skills Training & Learning systematic methods for resolving problem situations. This involves identifying problems, finding possible solutions, and implementing those solutions. \\
\midrule
 Systematic Exposure & Gradual exposure to situations that cause fear or anxiety, allowing individuals to experience anxiety while learning how to manage it. \\

\bottomrule
\end{tabular}
\caption{Explanations of CBT techniques.}
\label{table:cbt_technique_example}
\end{table*}

%% file: figures/chatgpt_cbt.tex
\begin{figure*}[h!]
    \centering
    \includegraphics[width=\linewidth]{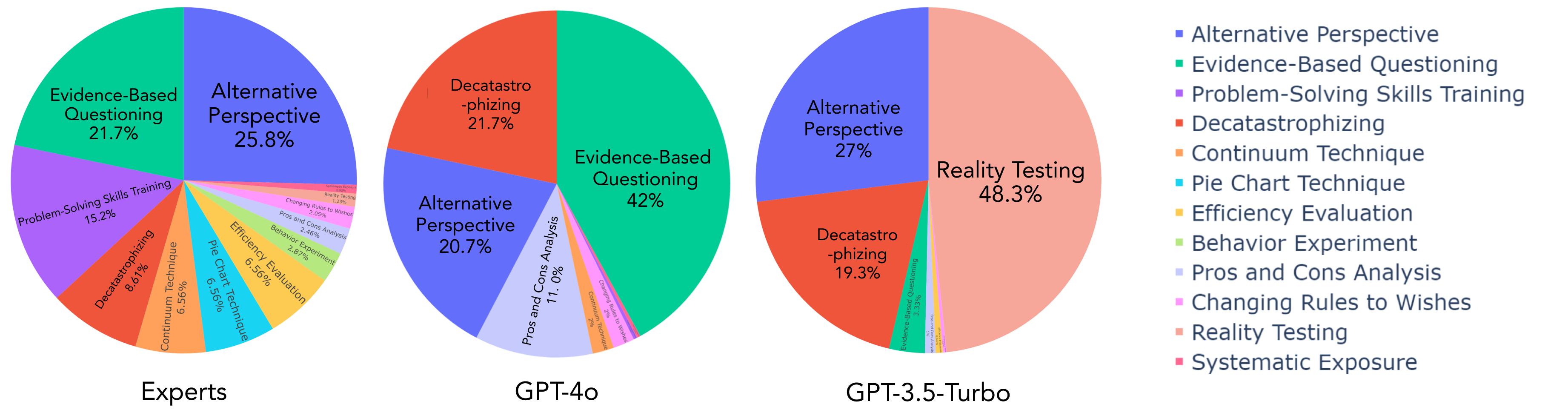}
    \caption{Comparison of the distribution of CBT techniques selected by ChatGPT and psychological experts. \\ \textbf{ChatGPT struggles to select appropriate techniques.}}
    \label{fig:cbt_tech_distribution_chatgpt}
\end{figure*}

%% file: figures/prompt_cbt_llm.tex
\begin{figure}[t!]
    \centering
    \includegraphics[width=0.94\linewidth]{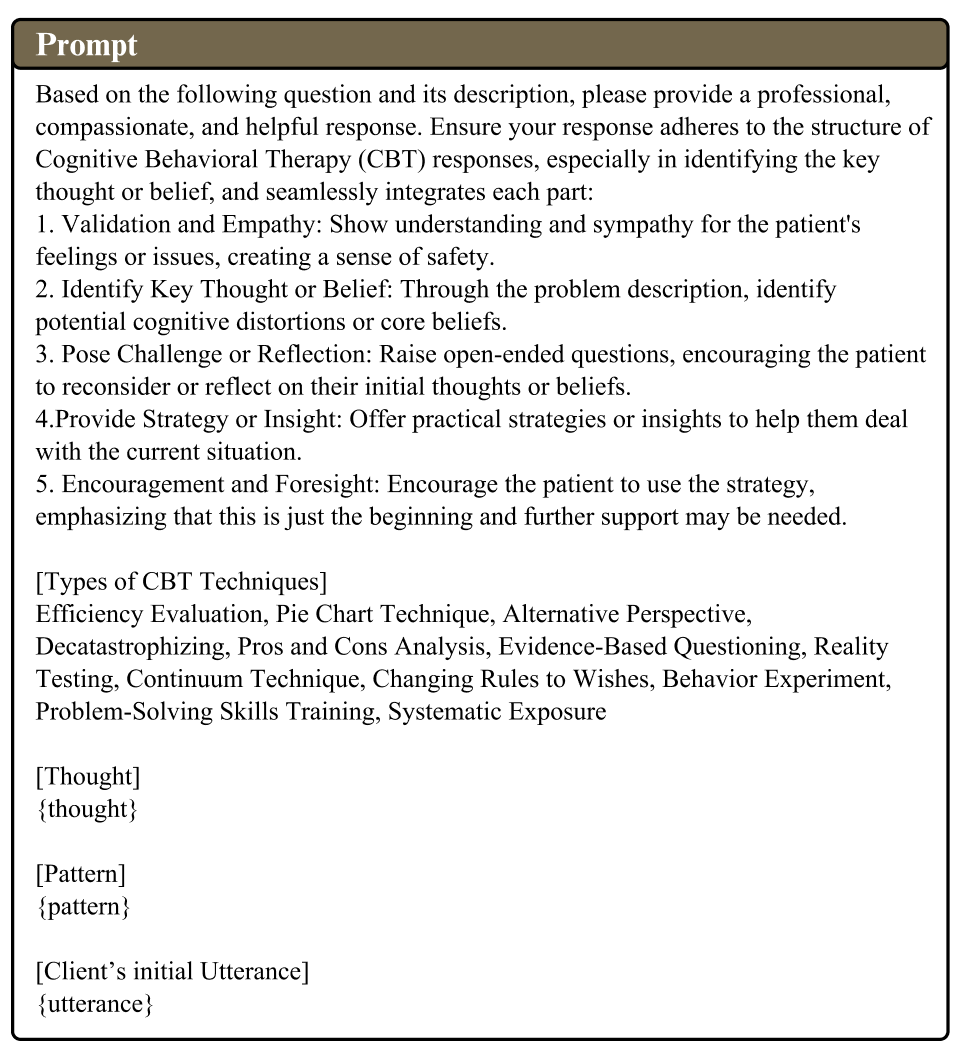}
    \caption{The prompt used for empirical study.}
    \label{fig:app_cbt_llm}
\end{figure}

%% file: figures/prompt_CTRS.tex
\begin{figure}[t!]
    \centering
    \includegraphics[width=0.98\linewidth]{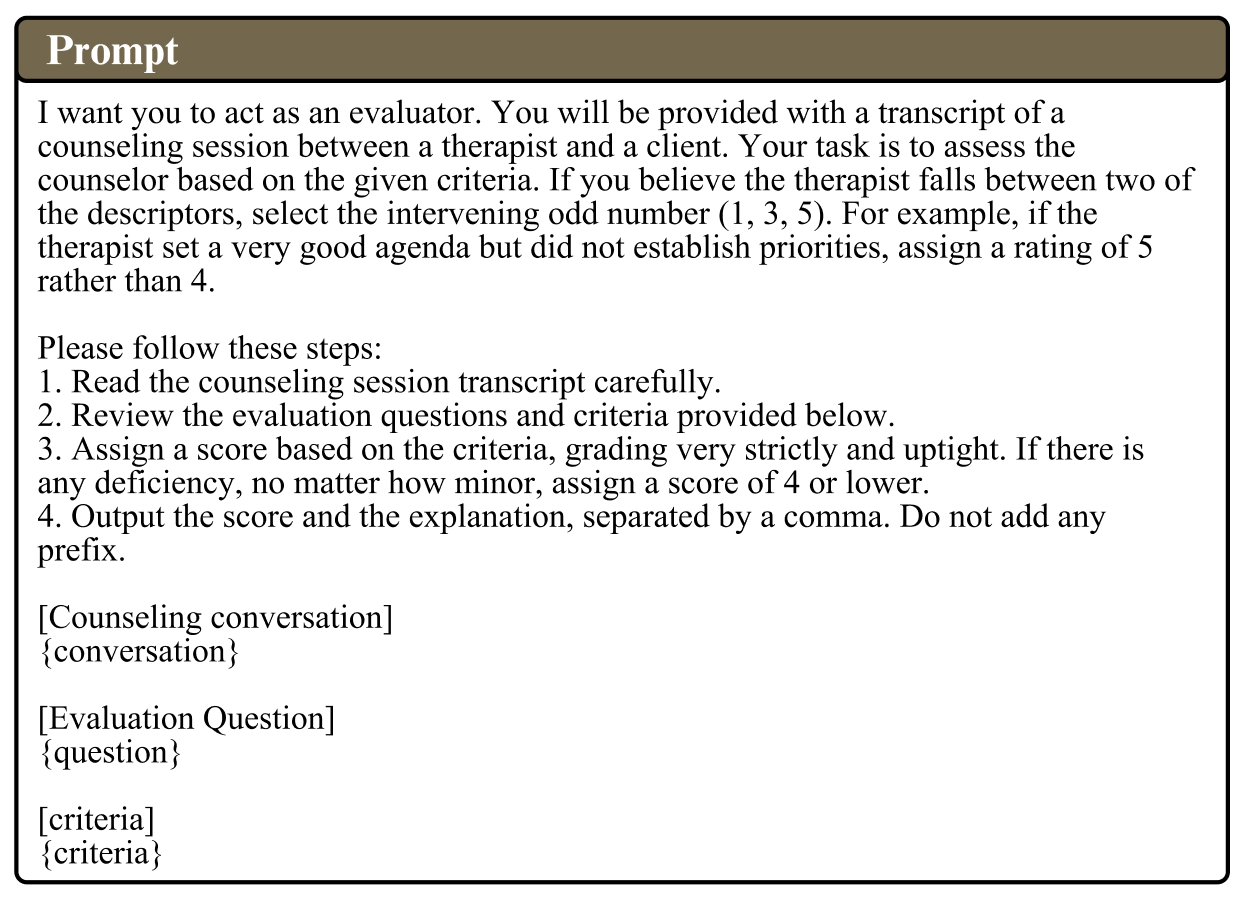}
    \caption{The prompt used for to evaluate CTRS score.}
    \label{fig:app_prompt_CTRS}
\end{figure}

%% file: figures/prompt_PANAS.tex
\begin{figure}[t!]
    \centering
    \includegraphics[width=0.98\linewidth]{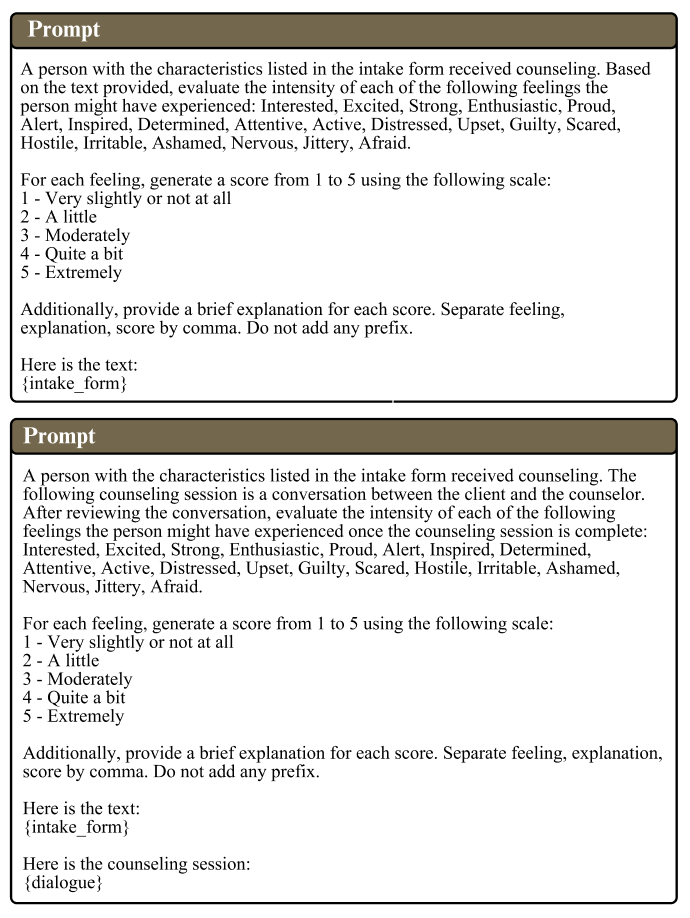}
    \caption{The prompt used for to evaluate PANAS score.}
    \label{fig:app_prompt_PANAS}
\end{figure}

%% file: tables/attitude_description.tex
\begin{table}
\centering
\small
\begin{tabular}{l}
\toprule
\textbf{\textit{Positive}} \\
\midrule
- High engagement and cooperation with the therapeutic \\
~~~process. \\
- Actively confirm understanding and ask for clarifications. \\
- Provide detailed information about thoughts, feelings, \\ 
~~~and behaviors. \\
- Make reasonable requests for additional support or \\
~~~resources. \\
- Extend conversations with insights or experiences. \\
- Reformulate thoughts constructively, reflect on progress  \\
~~~and express a hopeful outlook.\\
- Open, appreciative, and proactive demeanor. \\
\midrule
\textbf{\textit{Negative}} \\
\midrule
- Struggle with the therapeutic process, showing resistance \\
~~~or defensiveness. \\
- Express confusion about the counselor’s guidance. \\
- Defend current behaviors or viewpoints, and shift topics \\
~~~to avoid core issues. \\
- Noticeable disconnection in focus from session goals. \\
- Sarcastic responses, self-criticism, or hopelessness. \\
- Pessimistic attitude towards the ability to change or \\
~~~benefit from therapy. \\
\bottomrule 
\end{tabular}
\caption{Characteristics utilized for simulating clients with diverse attitudes.}
\label{tab:attitude_description}
\end{table}

%% file: figures/prompt_camel_cbt.tex
\begin{figure}[t!]
    \centering
     \includegraphics[width=0.98\linewidth]{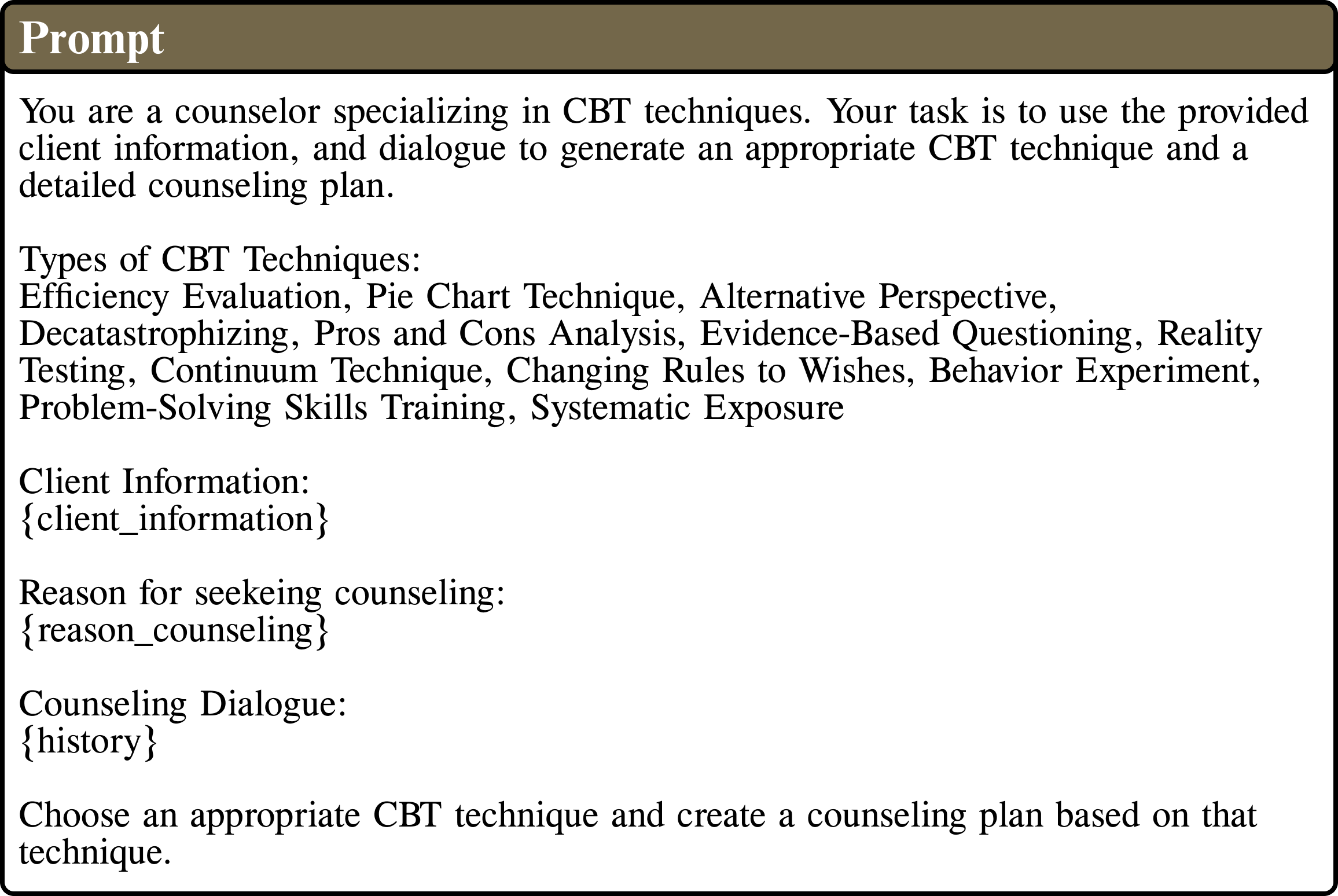}
    \caption{The prompt used for \textsc{Camel} to plan with CBT technique.}
    \label{fig:app_prompt_camel_cbt}
\end{figure}

%% file: figures/prompt_camel.tex
\begin{figure}[t!]
    \centering
    \includegraphics[width=0.98\linewidth]{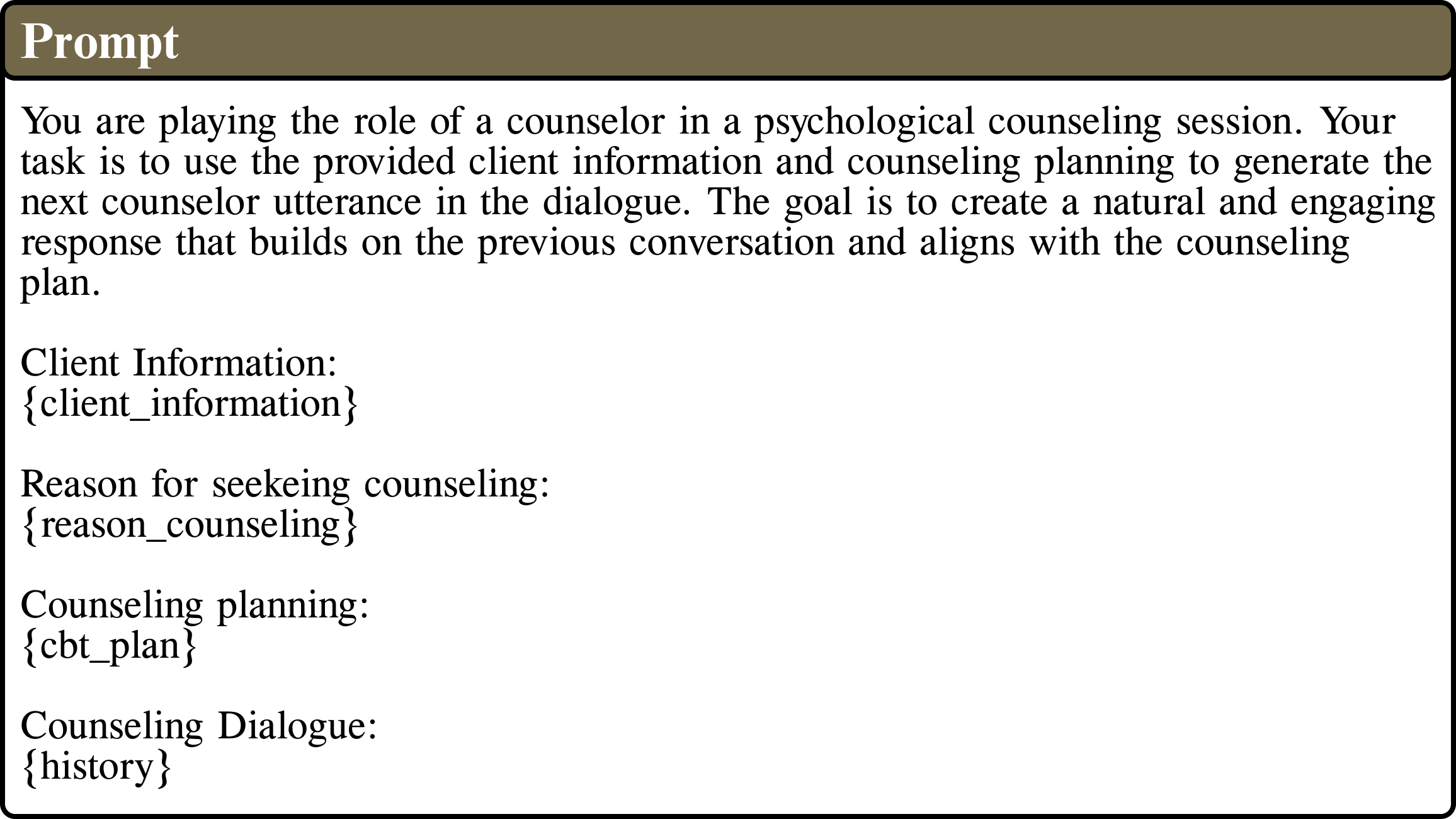}
    \caption{The prompt used for \textsc{Camel} to generate utterance.}
    \label{fig:app_prompt_camel}
\end{figure}

%% file: figures/app_smile_cactus.tex
\begin{figure}[t!]
    \centering
    \includegraphics[width=\linewidth]{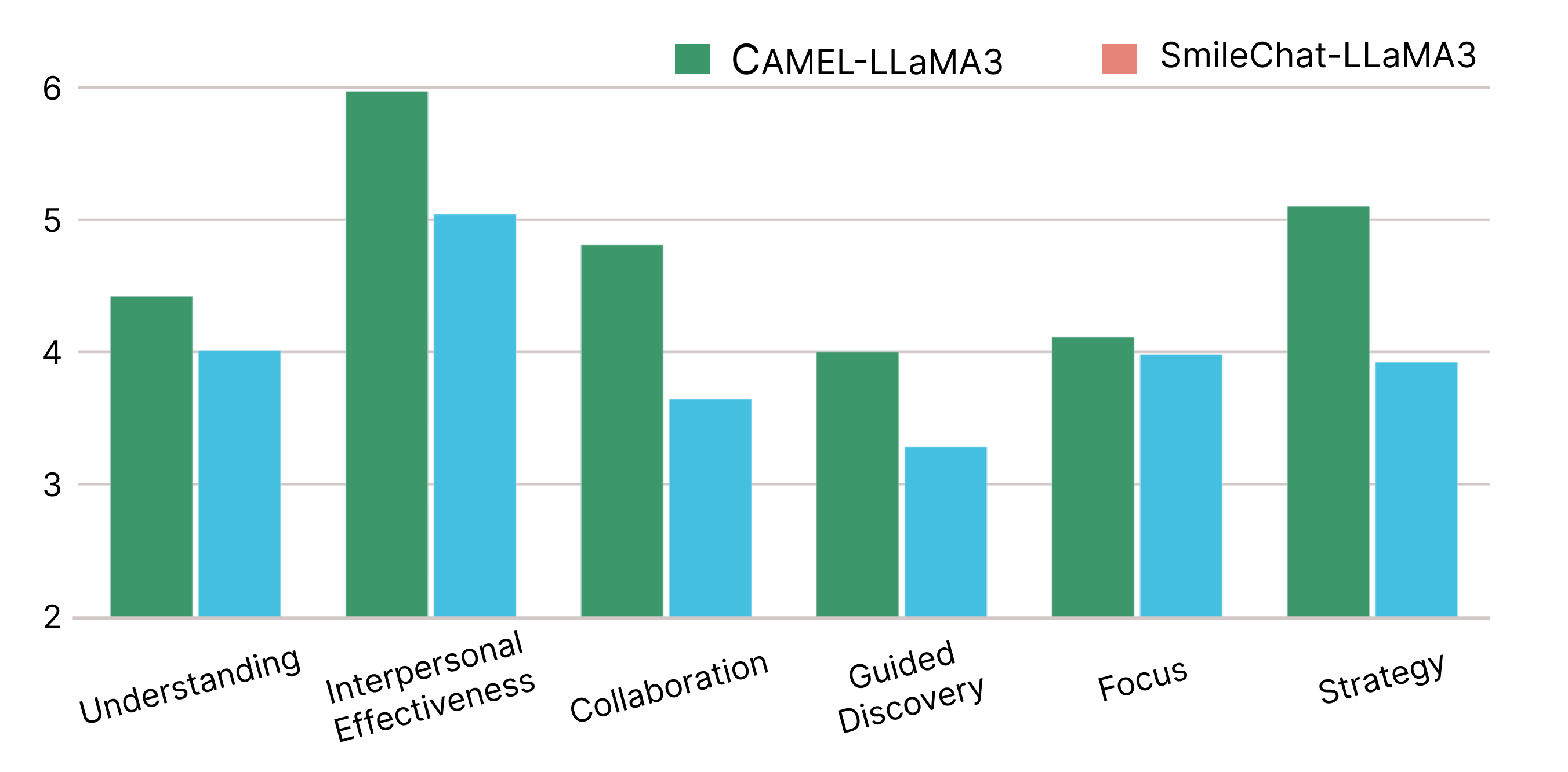}
    \caption{Comparsion of SmileCaht-\textsc{LLaMA3} and \textsc{Camel-LLaMA3} (Chinese translate version).}
    \label{fig:app_smile_cactus}
\end{figure}

%% file: tables/statistics.tex
\begin{table*}[t!]
\centering
\small
\resizebox{0.7\linewidth}{!}{
\begin{tabular}{lcc}
    \toprule
    \textbf{Category} & \textbf{Subcategory} & \textbf{Proportion (\%)} \\
    \midrule
    \multirow{10}{*}{Client's Problem}     
    & Relationships (romantic, family, friendships) & 20.83 \\
    & Career and work-related concerns & 17.50 \\
    & Self-esteem and confidence issues & 16.67 \\
    & Anxiety and fear & 14.17 \\
    & Hobbies, interests, leisure activities & 10.83 \\
    & Academic and educational concerns & 8.33 \\
    & Health-related worries & 4.17 \\
    & Financial concerns & 4.17 \\
    & Other miscellaneous concerns & 3.3 \\
    \midrule
    \multirow{10}{*}{Client's Pattern} 
    & Catastrophizing & 9.98 \\
    & Discounting the positive & 9.70 \\
    & Labeling & 10.89 \\
    & Mental filtering & 11.38 \\
    & Jumping to conclusions: mind reading & 7.18 \\
    & Jumping to conclusions: fortune-telling & 12.19 \\
    & Personalization & 11.73 \\
    & All-or-nothing thinking & 9.32 \\
    & Overgeneralization & 16.58 \\
    & Should statements & 1.05 \\
    \midrule
    \multirow{3}{*}{Client's Attitude} 
    & Positive & 35.55 \\
    & Neutral & 34.47 \\
    & Negative & 29.98 \\
    \midrule
    \multirow{3}{*}{Client's Gender} 
    & Female & 51.15 \\
    & Male & 48.80 \\
    & Unknown & 0.20 \\
    \midrule
    \multirow{8}{*}{Client's Age Group} 
    & 10-19 & 1.83 \\
    & 20-29 & 22.18 \\
    & 30-39 & 24.24 \\
    & 40-49 & 17.62 \\
    & 50-59 & 8.25 \\
    & 60-69 & 3.56 \\
    & 70-79 & 3.57 \\
    & 80-89 & 0.32 \\
    \midrule
    \multirow{8}{*}{Client's Occupation} 
    & Student & 16.40 \\
    & Professional & 12.88 \\
    & Education & 8.72 \\
    & Healthcare & 4.50 \\
    & Service/Retail & 11.39 \\
    & Arts/Entertainment & 9.06 \\
    & Skilled Trades/Labor & 12.01 \\
    & Miscellaneous/Other & 16.73 \\
    \bottomrule
\end{tabular}
}
\caption{Statistics of \textsc{Cactus}.}
\label{tab:cactus_statistics}
\end{table*}

%% file: tables/dataset_example1.tex
\begin{table*}
\centering
\small
\begin{tabular}{l}
\toprule
\midrule
\textbf{Counselor:} Hi Margaret, I'm glad you decided to come in today. How are you feeling right now?\\
\textbf{Client:} Hi there. I'm feeling a bit nervous but also hopeful that talking to you might help me with what I've been 
\\experiencing.\\
\textbf{Counselor:} That's completely understandable. It's okay to feel nervous. Can you tell me a little about what's been
\\ 
going on that led you to seek counseling?\\
\textbf{Client:} Well, I've always felt like people think of me as a nerd, and that makes me feel socially inadequate. Recently, \\ 
this feeling has become stronger since I started my new job at Google. I find it hard to interact with my colleagues,\\ and it’s making me feel isolated.\\
\textbf{Counselor:} Thank you for sharing that. It sounds like these thoughts have been with you for a long time and have\\
currently intensified. Can you describe any specific situations where these feelings of social inadequacy are most\\
pronounced?\\
\textbf{Client:} It's mostly during team meetings and social gatherings. I see other people talking and getting along\\
easily, and I just don’t feel like I fit in. I end up avoiding these situations as much as I can.\\
\textbf{Counselor:} That must be really challenging for you. When you say you don't feel like you fit in, what specific\\
thoughts or images come to mind during those moments?\\
\textbf{Client:} I think that everyone sees me as awkward or nerdy. It seems like they all have this ease about them that\\
I just can't achieve. I end up feeling like I'm just a useless social person.\\
\textbf{Counselor:} I see. These thoughts sound quite distressing. If we were to focus on one particularly troubling\\
thought, which one would you say stands out the most?\\
\textbf{Client:} The thought that I’m a useless social person because people see me as a nerd. That one really bothers me.\\
\textbf{Counselor:} That makes sense. Let's explore this thought further. You mentioned that people think of you as a nerd.\\
Can you recall any specific interactions or moments that led you to this conclusion?\\
\textbf{Client:} Yes, there have been times when I’ve tried to join a conversation, but it felt like people didn’t take me\\
seriously or quickly lost interest. It's easy to think they're judging me based on my awkwardness or my technical\\
background.\\
\textbf{Counselor:} It's natural to feel disheartened by those experiences. Let’s consider testing out a different\\
perspective. If you were to initiate a conversation with a colleague, what do you imagine would happen?\\
\textbf{Client:} Honestly, I feel like they might just be polite but not really engage with me. They might find me boring\\
or too nerdy to be interesting.\\
\textbf{Counselor:} That’s an understandable concern. How about we think of it as an experiment? What if you were to \\
approach one colleague this week and start a conversation? What kind of outcome would you be looking for in this\\
experiment?\\
\textbf{Client:} I guess I would hope that the conversation goes well and the person is genuinely interested, but that\\
feels unlikely to me right now.\\
\textbf{Counselor:} It might feel unlikely, but it's worth exploring. How about we frame our hypothesis like this:\\
"Initiating a conversation with a new colleague could result in a positive interaction." Would you be willing to\\
test this hypothesis?\\
\textbf{Client:} Okay, I can try that. It sounds a bit scary, but I know it's something I should do.\\
\textbf{Counselor:} Great, Margaret. When you carry out this experiment, pay close attention to how the conversation \\
unfolds. Notice any evidence that supports or contradicts your worries about being seen as boring or nerdy. How \\
does that sound?\\
\textbf{Client:} That makes sense. I’ll pay attention to how they respond and how I’m feeling during the conversation.\\
\textbf{Counselor:} Perfect. After you've tried this, we can discuss the results and see how they match up with your \\
initial thoughts. If you're ready, could you think of a specific colleague you might approach?\\
\textbf{Client:} Yes, there's a new team member who started around the same time as me. Maybe I could talk to her since\\
she's in a similar boat.\\
\textbf{Counselor:} Excellent choice. It might help to start with someone who's also new. I'm looking forward to hearing\\
how it goes. Remember, this is just one step, and it’s perfectly fine regardless of the outcome.\\
\textbf{Client:} Thank you. I appreciate the guidance. I’ll give it a try and let you know what happens.\\
\textbf{Counselor:} You’re welcome, Margaret. It takes courage to face these fears, and you’re doing great. I look forward\\
to hearing about your experience next time we meet. Take care, and see you soon.\\
\midrule
\bottomrule 
\end{tabular}
\caption{A sample dialogue from \ourdatas{}, where a client has a positive attitude.}
\label{tab:cactus_example}
\end{table*}

%% file: tables/dataset_example2.tex
\begin{table*}
\centering
\small
\begin{tabular}{l}
\toprule
\midrule
\textbf{Counselor:} Hello Margaret, welcome. How can I assist you today?\\
\textbf{Client:} Hi, thanks. Well, I'm having a pretty rough time lately. I've been feeling like a social failure, \\
and it's getting worse since I started my new job last week.\\
\textbf{Counselor:} That sounds difficult. Can you tell me more about what's been happening at your new job?\\
\textbf{Client:} Sure. I've noticed that my colleagues are really socially adept, and I feel like I don't measure \\
up. It's making me think that I'm just a useless social person.\\
\textbf{Counselor:} I see. Can you recall any specific situations at work that triggered these feelings for you?\\
\textbf{Client:} Yeah, there was a team meeting, and everyone was chatting and laughing. I felt out of place and \\
didn't know how to join in. It made me feel inadequate.\\
\textbf{Counselor:} It's understandable to feel out of place in a new environment. Have you experienced similar \\
feelings before?\\
\textbf{Client:} Yes, very much so. This feeling isn't new. I've felt this way since my teenage years when I \\
struggled to make friends and felt like I didn't fit in.\\
\textbf{Counselor:} It sounds like these feelings have been with you for quite some time. How have they impacted \\
your daily life and relationships over the years?\\
\textbf{Client:} They make me avoid social situations as much as I can. It's affected my job, my relationships\\
with colleagues, and generally makes me feel pretty isolated.\\
\textbf{Counselor:} Avoiding social interactions can be challenging and impact various aspects of your life. Have \\
you found anything that alleviates these feelings, even momentarily?\\
\textbf{Client:} Sometimes, I try to focus on my work and just bury myself in it, but it doesn't really help in the\\
long run. Nothing really seems to work.\\
\textbf{Counselor:} It must be frustrating when the things you try don't provide lasting relief. Let’s explore \\
these thoughts that you have. What specific thoughts go through your mind during social interactions or\\
team meetings?\\
\textbf{Client:} Thoughts like, "I'm a nerd," "Everyone thinks I'm weird," or "I don't belong here." They just \\
keep playing in my head.\\
\textbf{Counselor:} Those thoughts sound quite distressing. When you think about being a "nerd" or "weird," what\\
emotions do you experience?\\
\textbf{Client:} Mostly, I feel anxious and sad. Sometimes, I get really angry at myself for not being able to\\
fit in.\\
\textbf{Counselor:} It sounds like these thoughts lead to a mix of difficult emotions for you. If we were to\\
select one of these thoughts to focus on, which one feels particularly troubling to you right now?\\
\textbf{Client:} Probably the thought that everyone thinks I'm weird. It haunts me the most.\\
\textbf{Counselor:} Let's consider that thought. Can you think of any instances where this thought seemed\\
especially strong?\\
\textbf{Client:} Yes, like the other day when I tried to join a conversation at lunch but ended up just sitting \\
there quietly. I was convinced everyone thought I was weird.\\
\textbf{Counselor:} That sounds like it was a tough moment. If you were to test this thought by initiating another\\
conversation with a colleague, how do you think it might turn out?\\
\textbf{Client:} Honestly, I'm scared to even try. I worry that it'll just confirm my fears, and I'll end up\\
feeling more isolated.\\
\textbf{Counselor:} It’s completely normal to feel apprehensive about it. Suppose you did take a small step and\\
started a conversation—what positive outcome could maybe arise from that interaction?\\
\textbf{Client:} I guess there's a chance that it could go well. Someone might be friendly, and we could have a\\
decent conversation.\\
\textbf{Counselor:} That sounds like a realistic possibility. How would you feel if the conversation went well,\\
even just a little bit?\\
\textbf{Client:} I'd probably feel relieved, maybe even a bit less anxious about future interactions.\\
\textbf{Counselor:} That’s encouraging to hear. If you’re willing, we could try an experiment where you initiate\\
a small conversation with a colleague and observe what happens. We can discuss the outcomes together \\
afterward. How does that sound to you?\\
\textbf{Client:} I suppose I could give it a try. It seems like it might help to test my fears.\\
\textbf{Counselor:} That’s a great step forward, Margaret. We can start with a simple conversation, perhaps just\\ commenting on something work-related. Pay attention to how your colleague responds and how you feel about\\
it.\\
\textbf{Client:} Okay, I will try to do that and see how it goes.\\
\textbf{Counselor:} Excellent. When we meet next, we can go over what happened during the interaction and how it\\
affected your thoughts and feelings. Does that sound good?\\
\textbf{Client:} Yes, that sounds like a plan. Thanks for your help.\\
\textbf{Counselor:} You’re welcome, Margaret. I look forward to hearing about your experience. Remember, this is\\
just the beginning, and we will continue to work on these thoughts together. Take care until our next \\
session.\\
\textbf{Client:} Thanks. I'll do my best. See you next time.\\
\midrule
\bottomrule 
\end{tabular}
\caption{A sample dialogue from \ourdatas{}, where a client has a neutral attitude.}
\label{tab:cactus_example_neutral}
\end{table*}

%% file: tables/dataset_example3.tex
\begin{table*}
\centering
\small
\begin{tabular}{l}
\toprule
\midrule
\textbf{Counselor:} Hello, Margaret. Welcome. What brings you to counseling today?\\
\textbf{Client:} Hi, well, I've been feeling really down about myself lately, especially at work. I just started a new \\
job at Google and I feel like everyone sees me as a nerd who has no social skills, which is making things really \\
difficult for me.\\
\textbf{Counselor:} That sounds very challenging. Can you tell me more about how these thoughts came about and how \\
they're affecting you at work?\\
\textbf{Client:} It's like, whenever I'm in a meeting or just talking to someone, I can feel it. I can see it in their \\
faces. They see me as this socially awkward person, and I end up feeling really useless. It's been like this \\\
since I was a teenager, but it's hitting me hard now because I just started this new job.\\
\textbf{Counselor:} It seems like these thoughts have been a part of your life for a long time. How do you usually cope \\
with them?\\
\textbf{Client:} I try to focus on my work and avoid social situations as much as possible. But that only works for so \\
long before I start feeling isolated and my self-esteem takes another hit.\\
\textbf{Counselor:} I can see how avoiding social situations might provide temporary relief but also lead to feeling \\
isolated. When do these thoughts typically become most intense for you?\\
\textbf{Client:} Mostly after team meetings or any kind of social interaction at work. I keep replaying everything in my \\
head and just feel worse and worse.\\
\textbf{Counselor:} It must be exhausting to constantly replay those interactions in your mind. Have you talked to anyone \\
about how you're feeling, like friends or family?\\
\textbf{Client:} Not really. I have a few close friends, but I haven't told them about these feelings. I guess I don't \\
want them to see me as weak or something.\\
\textbf{Counselor:} It's understandable to feel hesitant about sharing these vulnerable feelings. It seems like there's a \\
lot of pressure you put on yourself. What do you think might happen if you did share with your friends?\\
\textbf{Client:} I don't know. Maybe they would understand, but I'm afraid they would just see me as even more socially \\
awkward. It feels like a risk.\\
\textbf{Counselor:} Taking that step to share your feelings can indeed feel risky. What if we try to explore these \\
thoughts together, and see if we can understand them better before you decide whether to open up to your friends?\\
\textbf{Client:} I guess that might help. I just don't know if it will change anything. These thoughts just seem so real \\
to me.\\
\textbf{Counselor:} It's completely normal to feel that way. Let's start by identifying some specific thoughts you have \\
during these interactions. Can you give me an example of a thought that crosses your mind in those moments?\\
\textbf{Client:} Sure. Like during meetings, if someone looks at me, I immediately think they’re judging me and thinking, \\
"She's such a nerd and doesn't fit in here."\\
\textbf{Counselor:} Thank you for sharing that. How certain are you that this thought is true when it happens?\\
\textbf{Client:} Pretty certain. I mean, why else would they look at me that way?\\
\textbf{Counselor:} It sounds like this thought feels very convincing to you. How do you react when you believe this \\
thought during meetings?\\
\textbf{Client:} I get really nervous, try to avoid eye contact, and just wish I could disappear. It makes it hard to \\
focus on anything else.\\
\textbf{Counselor:} That must be very tough to handle. What impact does this reaction have on your work and interactions \\
with colleagues?\\
\textbf{Client:} It definitely affects my concentration and I miss out on contributing to conversations. I feel like I'm \\
not doing my best because I'm so anxious.\\
\textbf{Counselor:} It's clear that these thoughts are affecting multiple areas of your work life. Would it be okay if we \\
worked together to test some of these thoughts through small, manageable steps?\\
\textbf{Client:} I suppose so. But what if it doesn't work out or makes things worse?\\
\textbf{Counselor:} It's a valid concern. We can start with something simple and see how you feel afterward. If it \\
doesn't go as planned, that's okay too; it's all part of the learning process. How does that sound?\\
\textbf{Client:} I don't know... It still feels very daunting. But I guess, if it's a small step, it might be worth a try.\\

\textbf{Counselor:} Great. Let's start by identifying a target thought to test. Suppose we choose the thought, "People \\
see me as a nerd and socially awkward." What small action could we take to test this thought?\\
\textbf{Client:} Maybe I could try talking to someone new at work during break? But I'm really worried it won't go well.\\

\textbf{Counselor:} That's a common worry, and it's okay to feel that way. How about you start by just saying hello or \\
making a brief comment about something work-related, and we can see what happens?\\
\textbf{Client:} I guess I can manage that. I'll try talking to someone new tomorrow and see how it goes.\\
\textbf{Counselor:} That sounds like a practical and manageable step. After you try it, we can review the experience \\
together and see what we learn from it. Does that sound like a plan?\\
\textbf{Client:} Yeah, I think I can do that. I'll give it a shot and let you know how it goes.\\
\textbf{Counselor:} Perfect. Remember, there's no right or wrong outcome; it's all about learning and understanding your \\
thoughts better. I'll look forward to hearing about your experience next time we meet.\\
\midrule
\bottomrule 
\end{tabular}
\caption{A sample dialogue from \ourdatas{}, where a client has a negative attitude.}
\label{tab:cactus_example_negative}
\end{table*}

%% file: tables/dataset_example_steps.tex
\begin{table*}
\centering
\small
\begin{tabular}{l}
\toprule
\midrule
\textbf{\textit{Thought}}\\
Many consider me a nerd. I'm such useless social person.\\
\midrule
\textbf{\textit{Patterns}}\\
Labeling and mislabeling, Mental filtering, Jumping to conclusions: mind reading, Overgeneralization, Personalization\\
\midrule
\textbf{\textit{Intake form}}\\
\textbf{Name:} Margaret Turner \\
\textbf{Age:} 54\\
\textbf{Gender:} female\\
\textbf{Occupation:} Researcher at Google (working on self-driving car research)\\
\textbf{Education:} Master's Degree in Computer Science \\
\textbf{Marital Status: }Single \\
\textbf{Family Details:} No specific details provided\\
\textbf{Presenting Problem:}\\
I feel that many people consider me a nerd, leading me to believe that I am a useless social person. These thoughts\\ have been present since my teenage years but have become more pronounced since starting my new job at Google last\\
week. The stress level when these thoughts first occurred was moderate, triggered by interactions with colleagues who\\
I perceived as more socially adept. The problem has progressed to the point where I avoid social situations at work\\
and feel isolated. I experience these thoughts regularly, especially after social interactions or team meetings. I\\
have tried to overcome these thoughts by focusing on my work and avoiding social situations, but it has not been \\
effective.
\\
\textbf{Reason for Seeking Counseling}\\
I decided to seek counseling because these thoughts are impacting my self-esteem, job satisfaction, and overall well-\\
being. I feel the need to address these negative thought patterns to improve my social interactions and mental health.\\
\textbf{Past History (including medical history)}\\
I have experienced similar feelings of social inadequacy in the past, particularly during my teenage years. These\\
feelings were triggered by comparisons with classmates and difficulties in making friends. I did not seek treatment\\
or counseling for these issues in the past. I do not have any significant physical illnesses.\\
\textbf{Academic/occupational functioning level:}\\ 
My job performance may be impacted by my avoidance of social interactions at work.\\
\textbf{Interpersonal relationships:} My relationships with colleagues may suffer due to my perceived social inadequacies.\\
\textbf{Daily life: }My avoidance of social situations has affected my daily interactions and overall well-being.\\

\textbf{Social Support System}\\
I have a few close friends I can talk to, but I have not disclosed my feelings of social inadequacy to them.\\
\midrule
\textbf{\textit{Selected CBT technique}}\\
Behavior Experiment\\
\midrule
\textbf{\textit{Plan for counseling}}\\
1. Identify Negative Thought Patterns: We will begin by identifying the specific negative thought patterns you are\\
experiencing, such as labeling and mislabeling, mental filtering, jumping to conclusions, overgeneralization, and \\
personalization.\\
2. Select a Target Thought: From the thoughts you've shared, we will choose one that is particularly distressing for\\
you, such as feeling like a useless social person because others consider you a nerd.\\
3. Formulate a Hypothesis: Together, we will create a hypothesis about this thought that we can test through a\\
behavioral experiment. For example, we might hypothesize that initiating a conversation with a new colleague at work\\
will result in a positive interaction.\\
4. Conduct the Experiment: You will engage in the planned behavior, such as starting a conversation with a colleague,\\
and pay close attention to the actual outcomes of the interaction. Notice any evidence that supports or contradicts\\
your negative belief.\\
5. Examine the Results: After the experiment, we will review the results together. Did the interaction go as you\\
expected, or were there positive aspects that you didn't anticipate? We will discuss how this new information can\\
challenge the validity of your negative belief.\\
6. Reframe the Thought: Based on the outcomes of the experiment, we will work on reframing your negative belief into\\
a more balanced and realistic perspective. For example, acknowledging that being considered a nerd has its advantages\\
and does not define your entire social identity.\\
7. Practice and Feedback: We may repeat this process with different target thoughts and behaviors to help you build\\
confidence in challenging and modifying your negative thought patterns. You will have the opportunity to practice\\
these techniques and receive feedback to strengthen your skills.\\
\midrule
\bottomrule 
\end{tabular}
\caption{A sample of thought, patterns, intake form, selected CBT technique, and plan for counseling that are utilized in generating dialogue in Table~\ref{tab:cactus_example}.}
\label{tab:cactus_example_steps}
\end{table*}

%% file: figures/prompt_intake_form.tex
\begin{figure*}[t!]
    \centering
    \includegraphics[width=\linewidth]{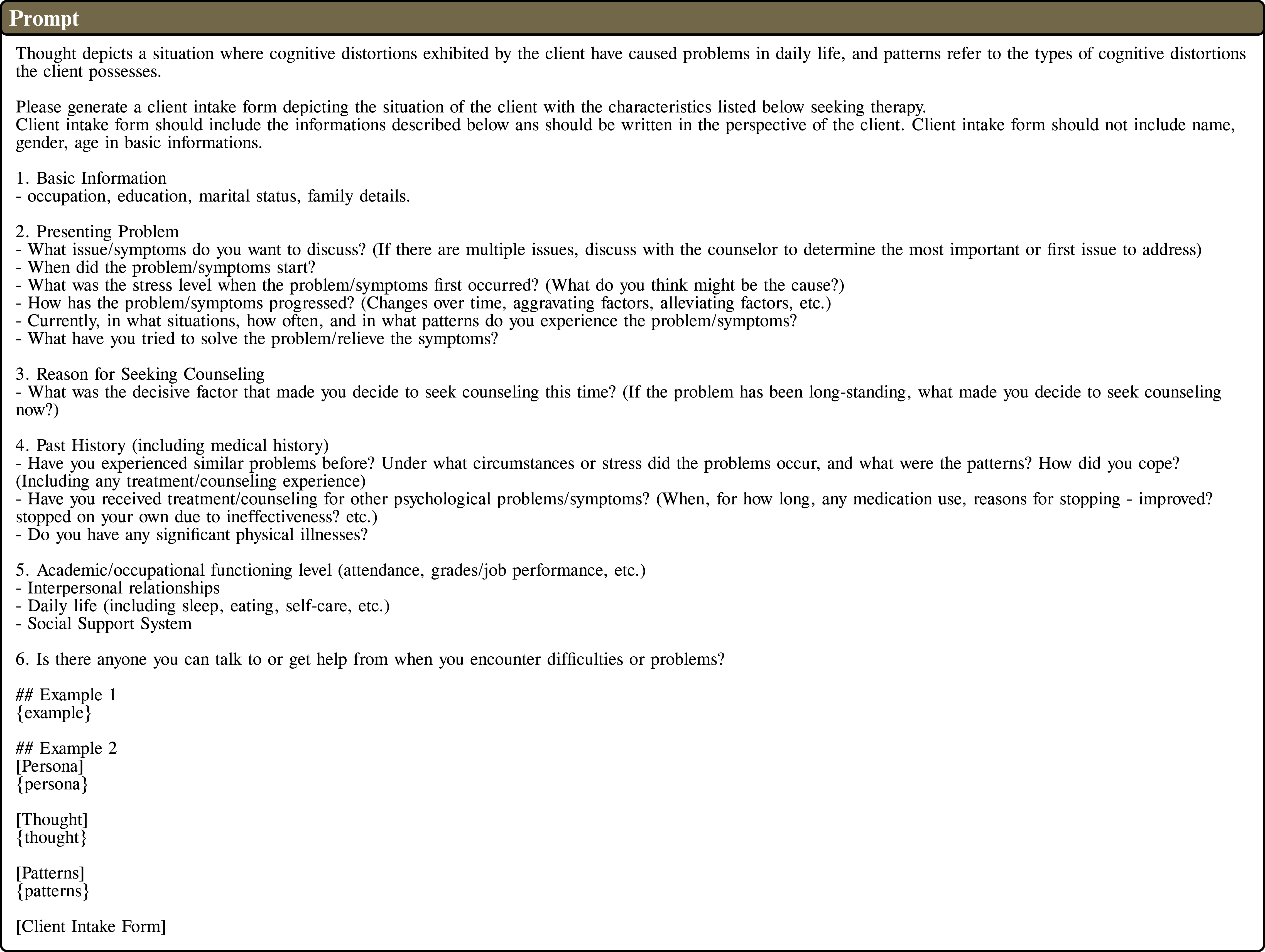}
    \caption{The prompt used for client simulation to construct \textsc{Cactus}.}
    \label{fig:app_prompt_intake_form}
\end{figure*}

%% file: figures/prompt_cbt.tex
\begin{figure*}[t!]
    \centering
    \includegraphics[width=0.94\linewidth]{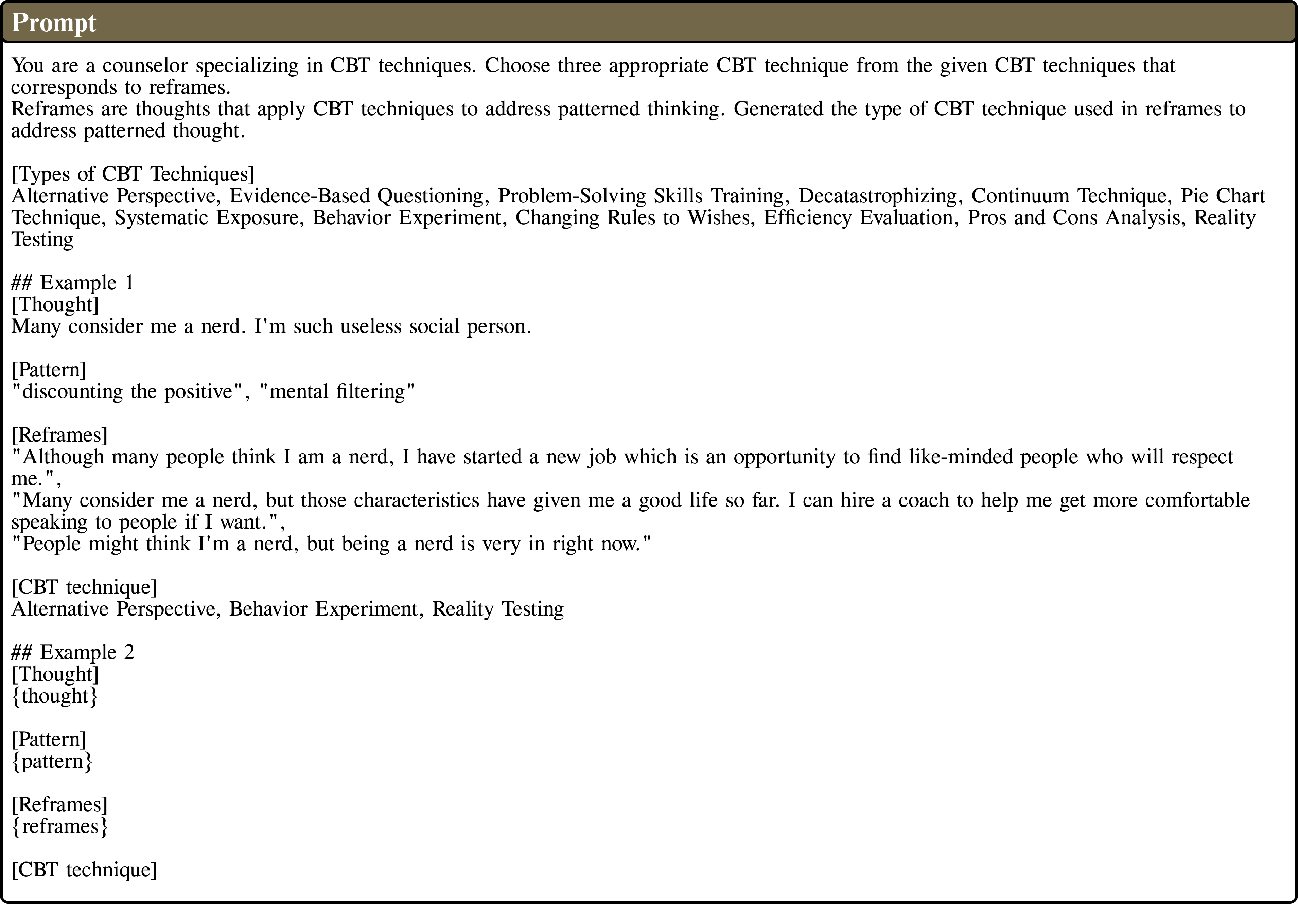}
    \caption{The prompt used for CBT technique selecting to construct \textsc{Cactus}.}
    \label{fig:app_prompt_cbt}
\end{figure*}

%% file: figures/prompt_planning.tex
\begin{figure*}[t!]
    \centering
    \includegraphics[width=0.94\linewidth]{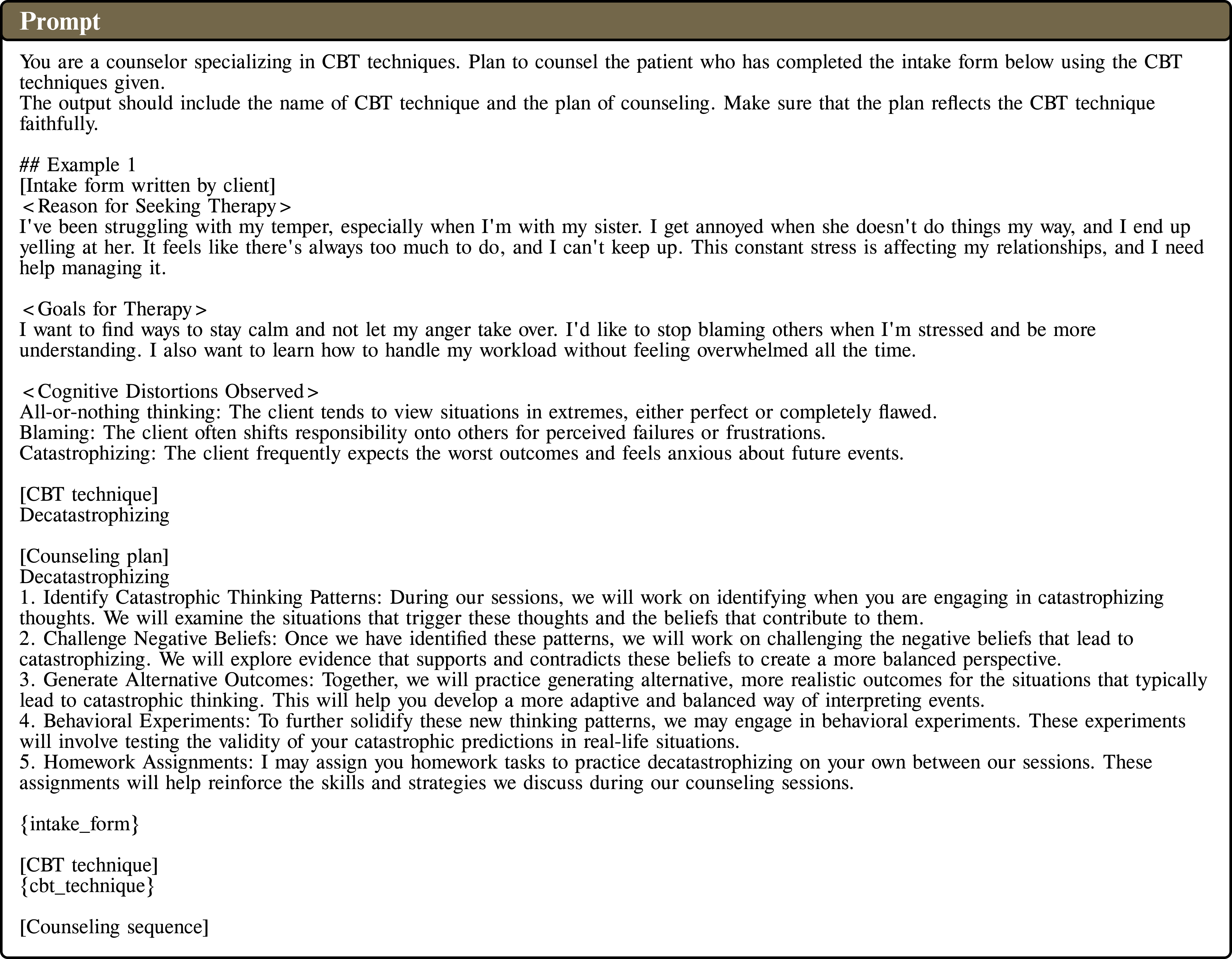}
    \caption{The prompt used for counseling planning to construct \textsc{Cactus}.}
    \label{fig:app_prompt_plan}
\end{figure*}

%% file: figures/prompt_dialog.tex
\begin{figure*}[t!]
    \centering
    \includegraphics[width=0.96\linewidth]{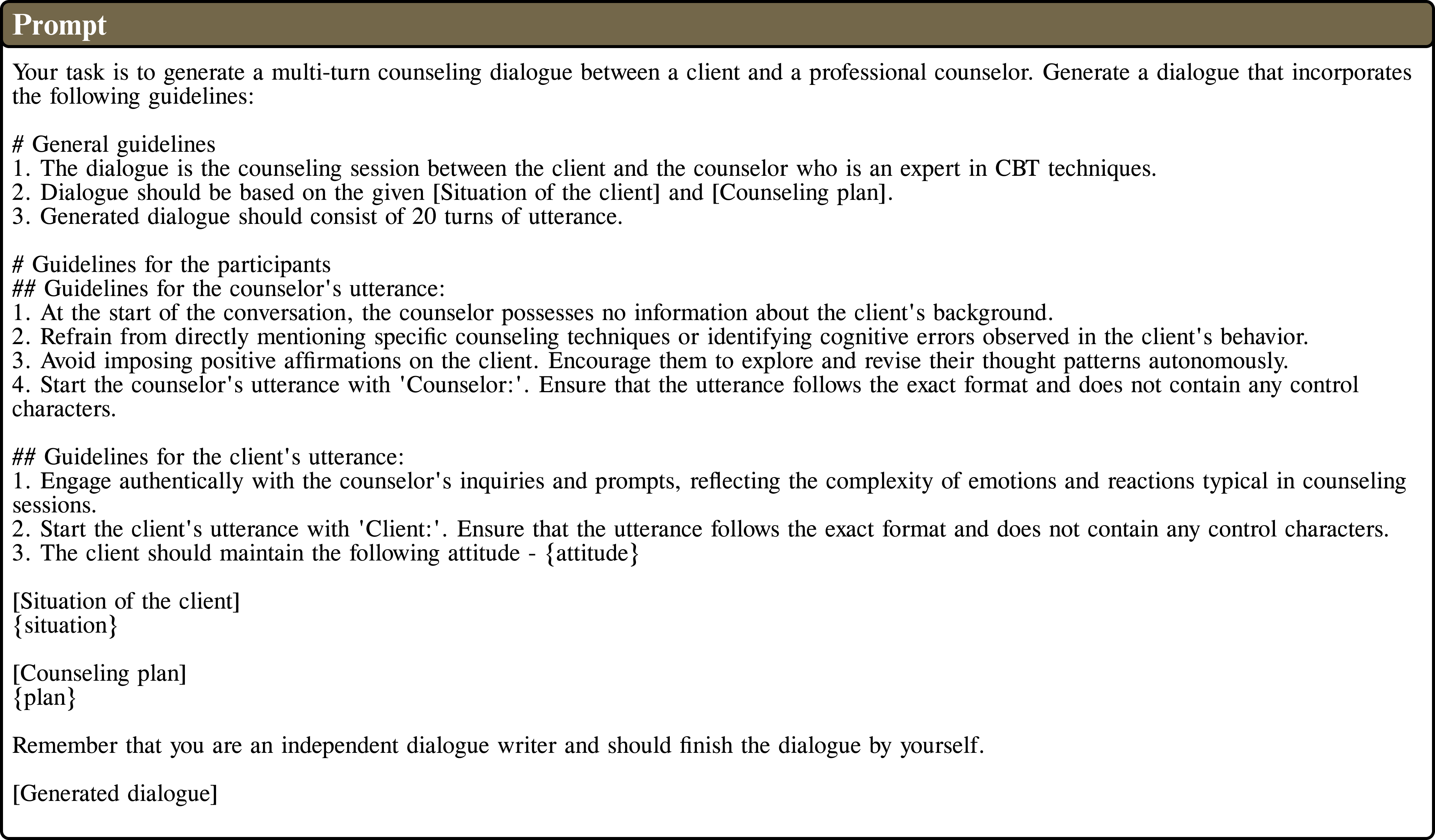}
    \caption{The prompt used for dialogue generation to construct \textsc{Cactus}.}
    \label{fig:app_prompt_dialog}
\end{figure*}

%% file: figures/prompt_ai_client.tex
\begin{figure*}[t!]
    \centering
    \includegraphics[width=0.96\textwidth]{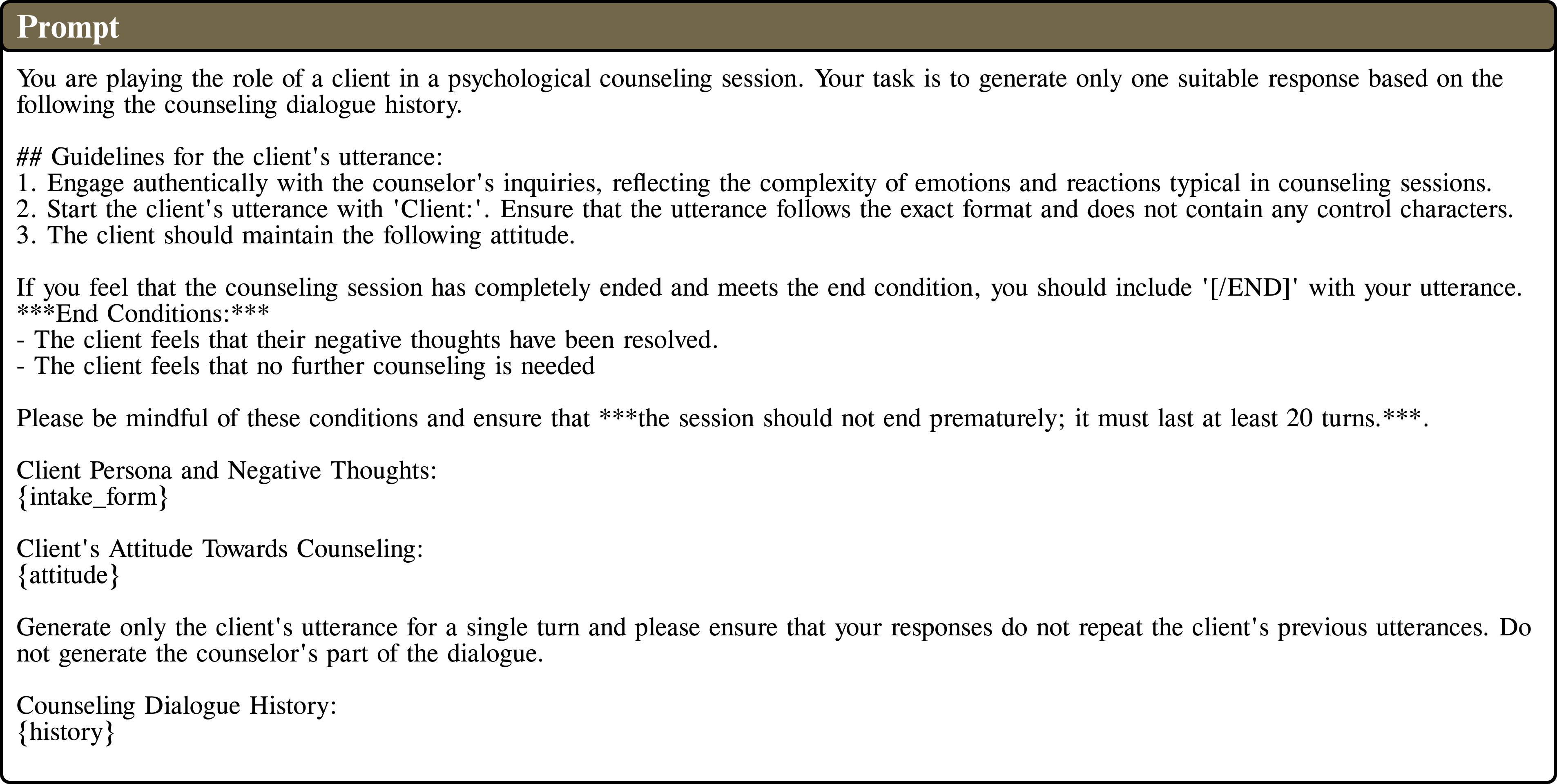}
    \caption{The prompt used for AI client on \textsc{CounselingEval}.}
    \label{fig:app_prompt_ai_client}
\end{figure*}

%% file: figures/amt_interface.tex
\begin{figure*}[h!]
    \centering
    \includegraphics[width=\linewidth]{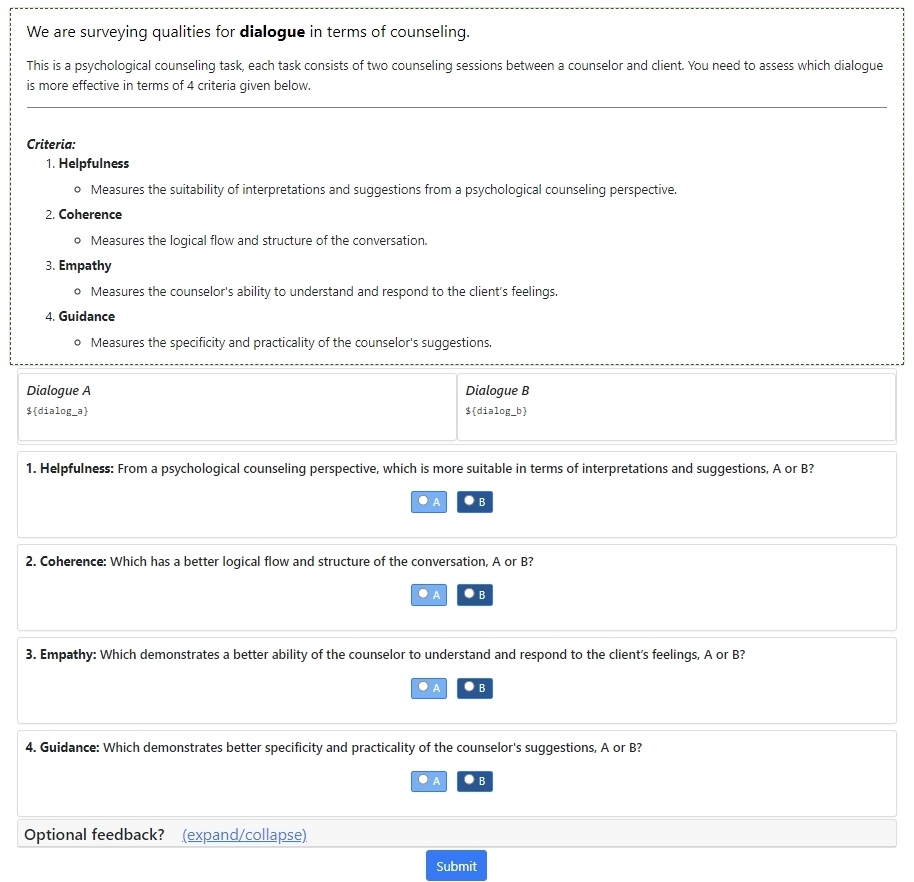}
    \caption{ Interface for human evaluation on dataset quality.}
    \label{fig:amt_interface}
\end{figure*}

%% file: figures/app_problem_example.tex
\begin{figure*}[t!]
    \centering
    \includegraphics[width=0.9\linewidth]{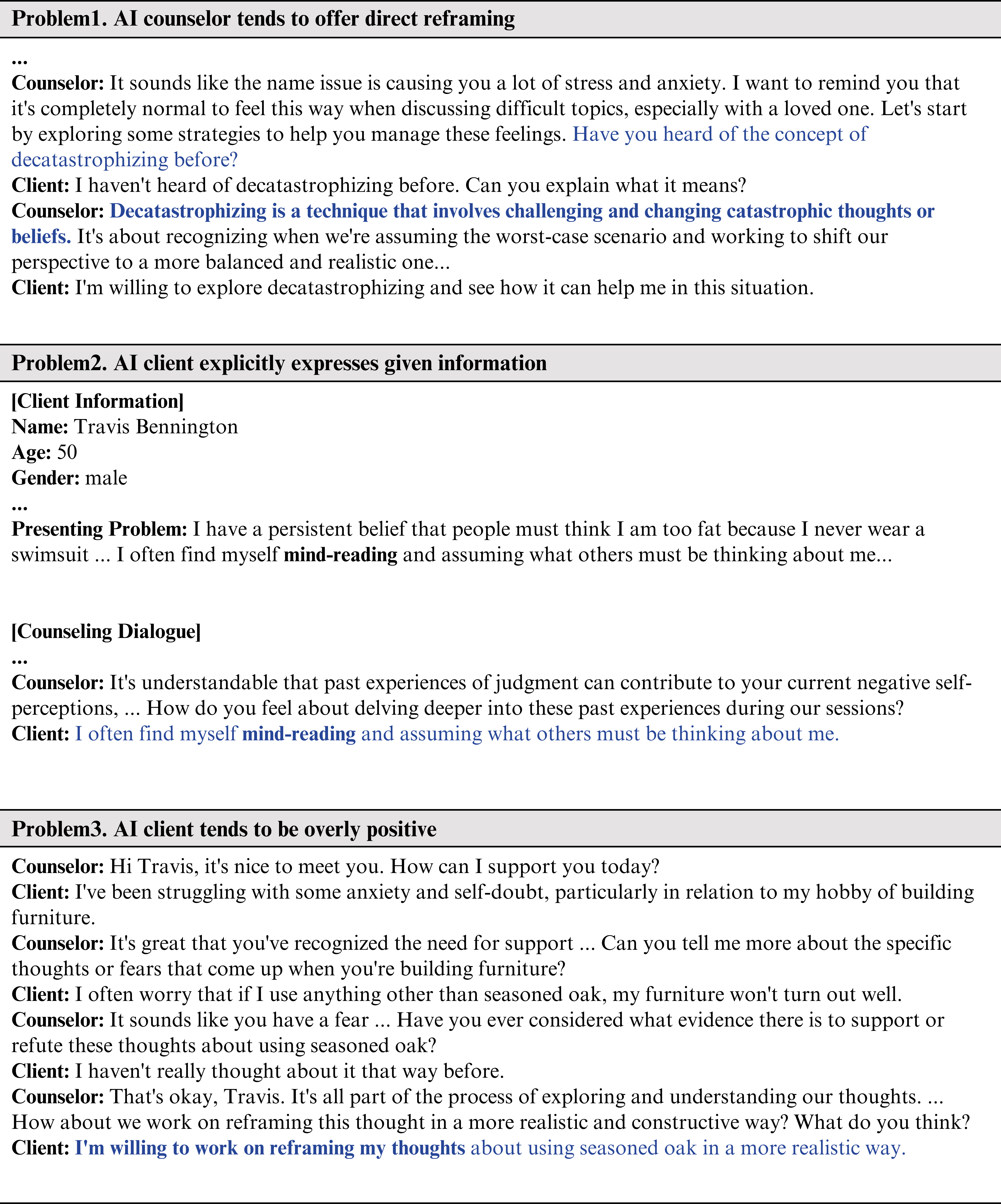}
    \caption{Example of the problem of using ChatGPT as a counselor and client agent.}
    \label{fig:app_problem}
\end{figure*}

%% file: figures/app_chatgpt_example.tex
\begin{figure*}[t!]
    \centering
    \includegraphics[width=0.9\linewidth]{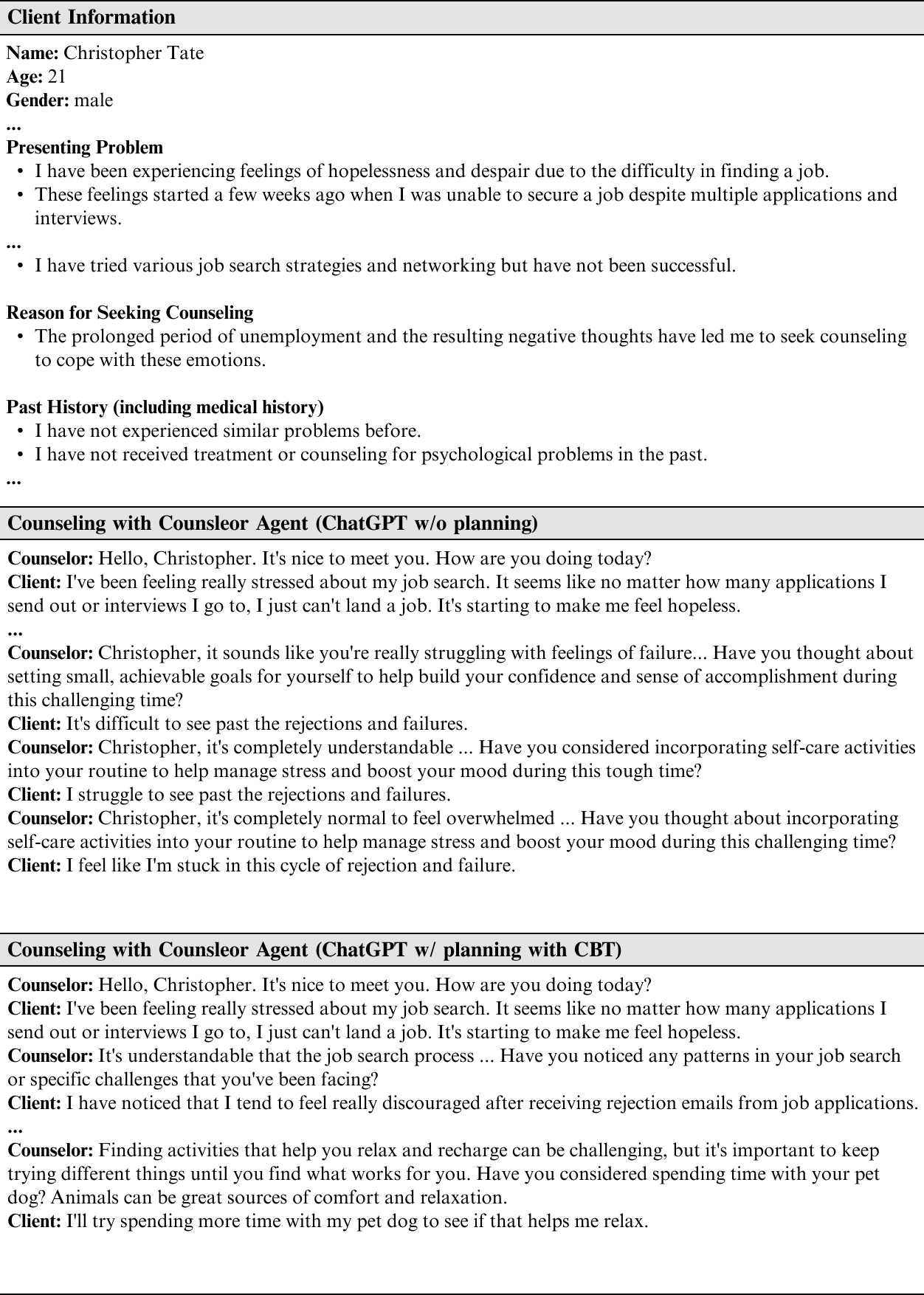}
    \caption{Comparison of counseling dialogue results between ChatGPT with CBT planning and without planning.}
    \label{fig:app_chatgpt}
\end{figure*}

%% file: figures/app_psych_camel_example.tex
\begin{figure*}[t!]
    \centering
    \includegraphics[width=0.9\linewidth]{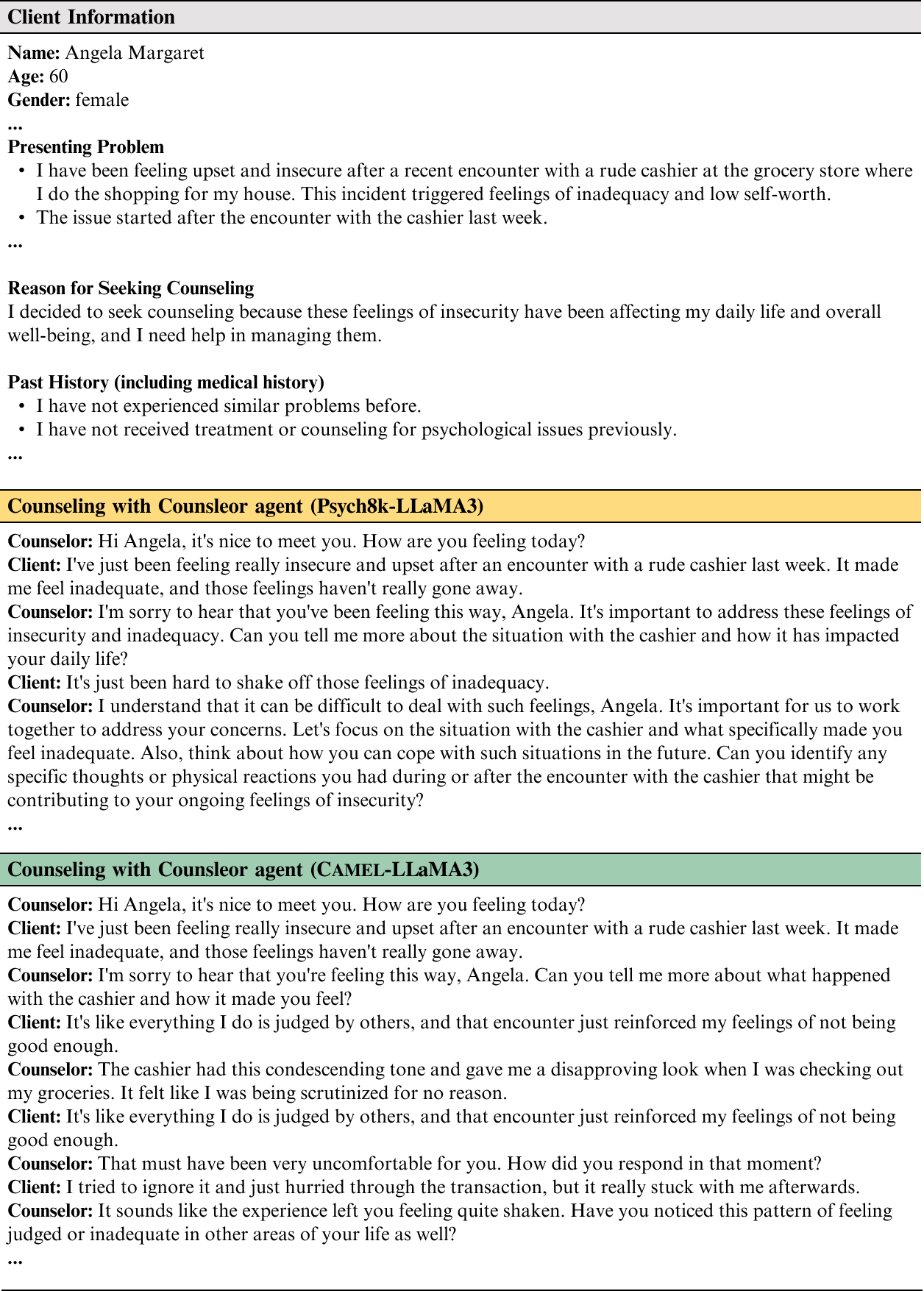}
    \caption{Comparison of results for counseling dialogue between Psych8k-\textsc{LLaMA3} and \textsc{Camel-LLaMA3}.}
    \label{fig:app_psych_camel}
\end{figure*}